\documentclass[10pt,journal,compsoc]{IEEEtran}

\usepackage{array}
\usepackage{bbm}
\usepackage{float}
\usepackage{units}
\usepackage{multirow}
\usepackage{color}
\usepackage{booktabs}
\usepackage{graphicx}
\usepackage{amsmath,amsfonts}

\newcommand{\reviewnew}[1]{{#1}}
\newcommand{\rreviewnew}[1]{{#1}}

\usepackage[hyphens,spaces,obeyspaces]{url}
\usepackage{hyperref}
\usepackage[T1]{fontenc}
\urldef{\footurl}\url{https://www.prophesee.ai/2023/02/27/prophesee-qualcomm-collaboration-snapdragon/}

\ifCLASSOPTIONcompsoc
  \usepackage[nocompress]{cite}
\else
  \usepackage{cite}
\fi

\usepackage[absolute]{textpos}
\begin{document}

% ============= Command to write a header to say "paper accepted at such conference"
\definecolor{somegray}{rgb}{0.5, 0.5, 0.5}
\newcommand{\darkgrayed}[1]{\textcolor{somegray}{#1}}
% The position is defined in absolute coords. Thus, if the page has another width,
% both numbers (4, 0.7) need to be adjusted. The argument {8} specifies the width of the textblock.
\begin{textblock}{10}(3, 0.25)
\begin{center}
\darkgrayed{This paper has been accepted for publication at the \\
IEEE Transactions on Pattern Analysis and Machine Intelligence, 2025. \textcopyright IEEE}
\end{center}
\end{textblock}
% ============= 


\title{Data-driven Feature Tracking for \\ Event Cameras with and without Frames}
% \title{Appendix: Data-driven Feature Tracking for \\ Event Cameras with and without Frames}


\author{Nico~Messikommer,~\IEEEmembership{Member,~IEEE,}
        Carter~Fang,~\IEEEmembership{Member,~IEEE,}
        Mathias~Gehrig,~\IEEEmembership{Member,~IEEE,}
        Giovanni~Cioffi,~\IEEEmembership{Member,~IEEE,}
        and~Davide~Scaramuzza,~\IEEEmembership{Senior Member,~IEEE}% <-this % stops a space
\IEEEcompsocitemizethanks{\IEEEcompsocthanksitem The authors are with the Robotics and Perception Group (\url{http://rpg.ifi.uzh.ch/}), University of Zurich, Switzerland.\protect\\
E-mail: nmessi@ifi.uzh.ch}% <-this % stops an unwanted space
% \thanks{Manuscript received April 19, 2005; revised August 26, 2015.}
}

\IEEEtitleabstractindextext{%
\begin{abstract}
Because of their high temporal resolution, increased resilience to motion blur, and very sparse output, event cameras have been shown to be ideal for low-latency and low-bandwidth feature tracking, even in challenging scenarios.
Existing feature tracking methods for event cameras are either handcrafted or derived from first principles but require extensive parameter tuning, are sensitive to noise, and do not generalize to different scenarios due to unmodeled effects. 
To tackle these deficiencies, we introduce the first data-driven feature tracker for event cameras, which leverages low-latency events to track features detected in an intensity frame.
We achieve robust performance via a novel frame attention module, which shares information across feature tracks.
Our tracker is designed to operate in two distinct configurations: solely with events or in a hybrid mode incorporating both events and frames. The hybrid model offers two setups: an aligned configuration where the event and frame cameras share the same viewpoint, and a hybrid stereo configuration where the event camera and the standard camera are positioned side-by-side. This side-by-side arrangement is particularly valuable as it provides depth information for each feature track, enhancing its utility in applications such as visual odometry and simultaneous localization and mapping.
\end{abstract}

}

\maketitle

% \IEEEdisplaynontitleabstractindextext

% \fi
\IEEEpeerreviewmaketitle

\IEEEraisesectionheading{\section*{Multimedia Material}}
\noindent A video is available at \url{https://youtu.be/dtkXvNXcWRY} and code at \url{https://github.com/uzh-rpg/deep_ev_tracker}
DOI: \url{https://doi.org/10.1109/TPAMI.2025.3536016}

\begin{figure}[t!]
\centering
\includegraphics[width=0.44\textwidth]{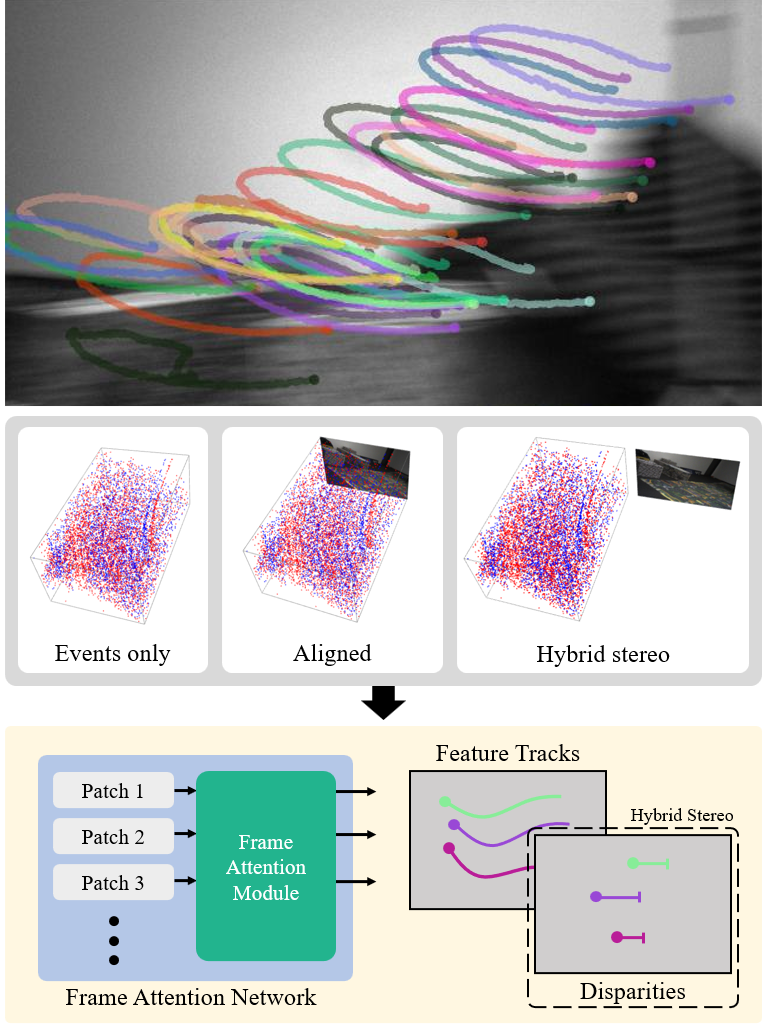}
\caption{
Our method leverages the high-temporal resolution of events to provide stable feature tracks in high-speed motion in which standard frames suffer from motion blur.
To achieve this, we propose a novel frame attention module that combines the information across feature tracks.
Our architecture seamlessly extends to sparse disparity estimation for a dual setup including a standard and event camera.
}
\label{fig:eye_catcher}
\end{figure} 

\section{Introduction}\label{sec:intro}
% \textbf{Why is the problem hard? What makes it challenging?}
\IEEEPARstart{D}{espite} many successful implementations in the real world, existing feature trackers are still primarily constrained by the hardware performance of standard cameras. 
To begin with, standard cameras suffer from a bandwidth-latency trade-off, which noticeably limits their performance under rapid movements:
at low frame rates, they have minimal bandwidth but at the expense of an increased latency; furthermore, low frame rates lead to large appearance changes between consecutive frames, significantly increasing the difficulty of tracking features.
At high frame rates, the latency is reduced at the expense of an increased bandwidth overhead and power consumption for downstream systems.
Another problem with standard cameras is motion blur, which is prominent in high-speed low-lit scenarios, see Fig.~\ref{fig:eye_catcher}.
These issues are becoming more prominent with the current commodification of AR/VR devices.  
Event cameras have been shown to be an ideal complement to standard cameras to address the bandwidth-latency trade-off~\cite{Gallego17pami,eklt}. 
Event cameras are bio-inspired vision sensors that asynchronously trigger information whenever the brightness change at an individual pixel exceeds a predefined threshold. 
Due to this unique working principle, event cameras output sparse event streams with a temporal resolution in the order of microseconds and feature a high-dynamic range and low power consumption. 
Since events are primarily triggered in correspondence of edges, event cameras present minimal bandwidth. 
This makes them ideal for overcoming the shortcomings of standard cameras.
% 

% 
% \textbf{B: How far has existing work come? What is the frontier?}
Existing feature trackers for event cameras have shown unprecedented results with respect to latency and tracking robustness in high-speed and high-dynamic range scenarios~\cite{eklt, Alzugaray2020HASTE:Events}.
Nonetheless, until now, event-based trackers have been developed based on classical model assumptions, which typically result in poor tracking performance in the presence of noise.
They either rely on iterative optimization of motion parameters~\cite{eklt, Kueng16iros, Zhu17icra} or employ a simple classification for possible translations of a feature~\cite{Alzugaray2020HASTE:Events}, thus, do not generalize to different scenarios due to unmodeled effects. 
% 
% \textbf{A: Why hasn’t the problem been solved? What is the stumbling block?}
% 
Moreover, they usually feature complex model parameters, requiring extensive manual hand-tuning to adapt to different event cameras and new scenes.
% 

% 
% \textbf{What does our paper contribute?}
To tackle these deficiencies, we propose the first data-driven feature tracker for event cameras, which leverages the high-temporal resolution of event cameras to maximize tracking performance.
Using a neural network, our method tracks features by localizing a template patch in subsequent event patches. 
The template patch can be constructed from a grayscale image either captured by a camera or generated by events using E2VID~\cite{Rebecq19cvpr}.
The network architecture features a correlation volume for the assignment and employs recurrent layers for long-term consistency.
To increase the tracking performance, we introduce a novel frame attention module, which shares information across feature tracks in one image.
We first train on a synthetic optical flow dataset and then finetune it with our novel self-supervision scheme based on 3D point triangulation using camera poses.
Moreover, our feature-tracking architecture seamlessly extends to dual-camera systems, where the event and the standard camera are placed side-by-side (a setup becoming popular in mobile device applications\footnote{\footurl}). 
This novel approach transforms the conventional challenge of aligning events and images into a strength since our method provides robust feature tracks with corresponding disparity information.
Both tasks are crucial for VO/SLAM pipelines relying on accurate feature tracking and disparity estimation.
As demonstrated in our experiments, the feature tracks and disparities inferred by our method provide more accurate estimations compared to existing methods, all while maintaining a fast runtime.
This manuscript extends our previous work~\cite{Messikommer23cvpr} in several ways:
\begin{itemize}
  \item We remove the requirement for aligned event and grayscale sensors by integrating our approach with an event-to-image reconstruction method and demonstrating competitive performance.
  \item We extend our work to sparse disparity estimation using dual-camera systems with minor modifications demonstrating the applicability of our method to VO/SLAM pipelines.
  \item Our experiments on sparse disparity estimation demonstrate that our approach performs comparably to dense disparity methods while requiring significantly less runtime.
\end{itemize}

\section{Related Work}
\label{sec:related_work}

\textbf{Frame-Based Feature Tracking} 
While no prior works have leveraged deep learning to track features from events, data-driven methods were recently proposed for feature tracking using standard frames. 
% 
% Persistent Independent Particles
Among them is PIP~\cite{harley2022pip}, which estimates the trajectories of queried feature locations for an entire image sequence and thus can even track features through occlusions by leveraging the trajectory before and after. 
% 
% Deep Patch Visual Odometry
Instead of processing the whole sequence, DPVO~\cite{dpvo} takes a sequence of images and simultaneously estimates scene depth and camera pose on-the-fly. 
It does so by randomly sampling patches from feature maps from frames and adding them to a bipartite frame graph, which is iteratively optimized by correlating feature descriptors from patches observed at different camera poses. 
A related research field to feature tracking is optical flow estimation, i.e., dense pixel correspondence estimation between two frames.
There exist many optical flow methods~\cite{flownet}, with correlation-based networks~\cite{flownet2, raft} being the state-of-the-art. 
However, despite recent advancements, frame-based feature trackers still suffer from the hardware limitation of standard cameras.
To tackle this disadvantage, we propose a self-supervised tracker that unlocks the robustness characteristics of event cameras for feature tracking and, by doing so, outperforms state-of-the-art tracking methods.

\textbf{Pose Supervision} 
Leveraging camera poses was previously explored for training feature detection and matching networks.
Wang \emph{et al.}~\cite{caps} used pose data to supervise a network for pixel-wise correspondence estimation where the epipolar constraint between two frames is used to penalize incorrect predictions. 
More recently, a correspondence refinement network called Patch2Pix~\cite{ZhouCVPRpatch2pix} extends the epipolar constraint supervision by using the Sampson distance instead of the Euclidean Distance.
Instead of only considering two camera poses, our self-supervision strategy computes a 3D point using DLT~\cite{abdel2015direct} for each predicted track in multiple frames, which makes our supervision signal more robust to errors.
Moreover, we supervise our network by computing a 2D distance between the reprojected and predicted points without the ambiguity of a distance to an epipolar line.
% 

% ========= Event-based Related Work ========= 
% 
\textbf{Event-Based Feature Tracking}
In recent years, multiple works have explored event-based feature tracking to increase robustness in challenging conditions, such as fast motion scenarios with large pixel displacement between timesteps and HDR scenes with very bright and dark areas~\cite{eklt}.
Early works~\cite{Ni12asynchronousevent-based, Kueng16iros, Zhu17icra} tracked features as point-sets of events and used ICP~\cite{besl15pami} to estimate the motion between timesteps, which can also be combined with frame-based trackers to improve performance~\cite{Dong21acm}.
Instead of point sets, EKLT~\cite{eklt} estimates the parametric transform between a template and a target patch of brightness increment images alongside the feature's velocity. 
Other event-based trackers align events along B\'{e}zier curves~\cite{Seok20WACV} or B-splines~\cite{Chui21Arxiv} in space and time to obtain feature trajectories.
To exploit the inherent asynchronicity of event streams, event-by-event trackers have also been proposed~\cite{Alzugaray18threedv, Dardelet2021arxiv}.
One of them is HASTE~\cite{Alzugaray2020HASTE:Events}, which reduces the space of possible transformations to a fixed number of rotations and translations. 
In HASTE, every new event leads to confidence updates for the hypotheses and a state transition if the confidence threshold is exceeded.
Another work called eCDT~\cite{Hu2022ECDT:Tracking} first represents features as event clusters and then incorporates incoming events into existing ones, resulting in updated centroids and, consequently, updated feature locations. 
In a similar direction to feature tracking, several event-based feature detectors~\cite{Chiberre21cvprw, Manderscheid19cvpr} were proposed, of which some are performing feature tracking based on proximity of detections in the image~\cite{Chiberre22arxiv, Alzugaray18ral}.
Apart from event-based feature tracking and detection, multiple works tackle the problem of object tracking using event cameras~\cite{Ramesh18bmvc, Chen19acm, Li19fn, Zhang21iccv, zhang2022cvpr, el2022ji}.

\begin{figure*}[ht!]
\centering
\includegraphics[width=1.0\textwidth]{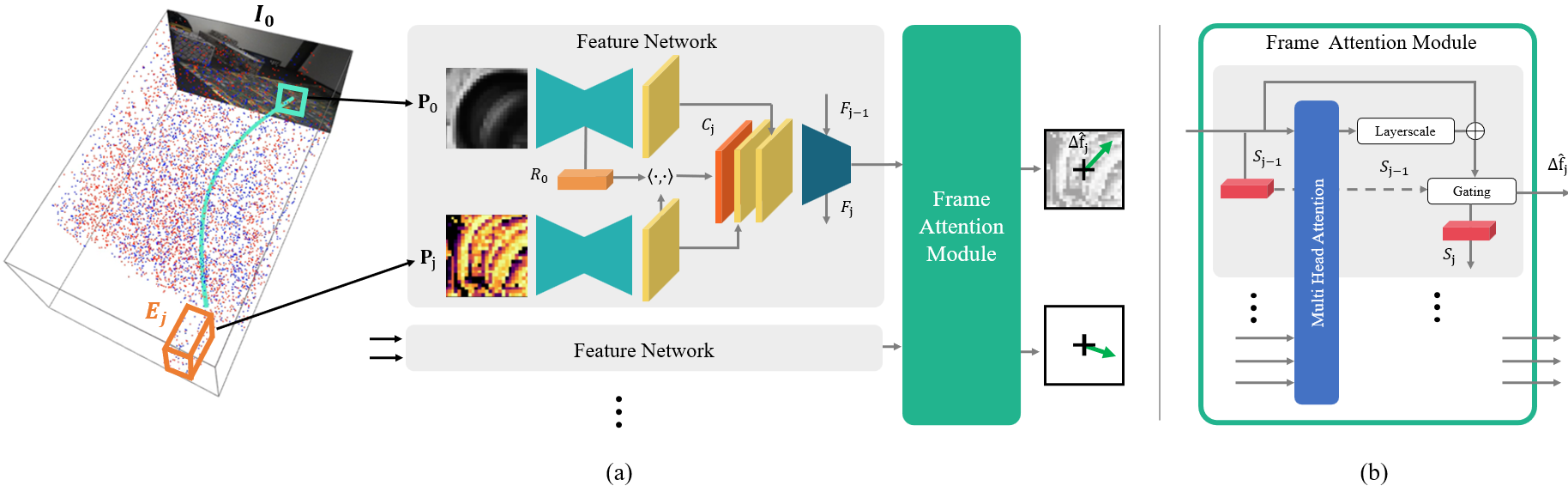}
\vspace*{-18pt}
\caption{
As shown in (a), our event tracker takes as input a reference patch $\mathbf{P_0}$ in a grayscale image $I_0$ and an event patch $\mathbf{P}_j$ constructed from an event stream $E_j$ at timestep $t_j$ and predicts the relative feature displacement $\mathbf{\Delta \hat{f}_{j}}$.
Each feature is individually processed by a feature network, which uses a ConvLSTM layer with state $F$ to process a correlation map $C_j$ based on a template feature vector $R_0$ and the pixel-wise feature maps of the event patch.
To share information across different feature tracks, our novel frame attention module (b) fuses the processed feature vectors for all tracks in an image using self-attention and a temporal state $S$, which is used to compute the final displacement $\mathbf{\Delta \hat{f}_{j}}$.
}
\vspace*{-12pt}
\label{fig:method}
\end{figure*} 
The task of optical flow estimation using event cameras gained popularity as well. 
Zhu \emph{et al.}~\cite{Zhu17icra} estimates the optical flow of features from events using ICP and an objective function based on expectation maximization to solve for the parameters of an affine transform.
More recently, an adaptive block matching algorithm~\cite{edflow} was proposed to estimate optical flow. 
Finally, recent data-driven methods for event-based optical flow estimation~\cite{evflownet, eraft, gehrig24tpami} leverage advances in deep optical-flow estimation.
Inspired by these advances, our tracking network leverages a correlation layer to update a feature's location. 

\textbf{Event-Based Depth Estimation}
Similar to template matching of events for feature tracking, the event streams of two side-by-side event cameras can also be used to compute the disparities by correspondence search.
Because of the high temporal resolution of event cameras, event-based stereo algorithms can leverage the time constraint in addition to the spatial similarity~\cite{steffen19neuro}, which is exploited in classical algorithms~\cite{kogler11advances, Rogister12nwls}.
Additionally, using Gabor filters to extract the orientation of edges, stereo event cameras can be used for 3D reconstruction ~\cite{camunas14frontier}.
The disparity map between two rectified event cameras can also be obtained with cooperative computing~\cite{firouzi16neural, Piatkowska2013AsynchronousSV}.
Leveraging deep learning architectures, stereo depth estimation approaches have been proposed that encode the event stream in a specialized event sequence embedding~\cite{Tulyakov19iccv}, or use a concentration network to focus event stacks~\cite{nam22cvpr}.
Beyond the depth estimation task, stereo event cameras were also employed in the context of SLAM~\cite{Zhou18eccv}.
Some methods were also proposed for disparity estimation using a hybrid event and frame camera setup.
They either rely on cross-correlation between binary edge and event frames combined with a completion network~\cite{wang2021iros}, a fully supervised disparity network~\cite{yifan21iros}, a self-supervised approach based on event-to-image reconstructions~\cite{gu2022jist}, or a two-stage matching network~\cite{zhang22eccv}.
Recent works have leveraged the relative camera pose to refine disparity estimation~\cite{chen24wacv} or combined it with a maximum shift distance method~\cite{kim23ral}.
Another direction is the depth prediction based on a monocular event camera using data-driven approaches ~\cite{RAL21Gehrig, Hidalgo20threedv}, which can also use self-supervision~\cite{Zhu19cvpr}, classical VO/SLAM algorithms~\cite{Kim2016RealTime3R, Rebecq17ral} or IMUs~\cite{Rebecq17bmvc, Zhu17cvpr}.
Recently, related work also explored the unique high temporal resolution of event cameras for active depth estimation~\cite{Muglikar213DV, Muglikar213DV_2, Brandli13fns, Matsuda15iccp, Martel18iscas, Mangalore20nfpp}.
In contrast to previous work, we compute the disparity only at selected features based on a standard and event camera in a side-by-side setting.
This requires only minor modifications to our proposed architecture for feature tracking, showcasing the modularity of our approach.

\section{Method}
\label{sec:method}
Feature tracking algorithms aim to track a given point in a reference frame in subsequent timesteps.
They usually do this by extracting appearance information around the feature location in the reference frame, which is then matched and localized in subsequent ones.
Following this pipeline, we extract an image patch $\mathbf{P}_0$ in a grayscale frame for the given feature location at timestep $t_0$ and track the feature using the asynchronous event stream.
The event stream $E_j=\{e_{i}\}_{i=1}^{n_j}$ between timesteps $t_{j-1}$ and $t_j$ consists of events $e_{i}$, each encoding the pixel coordinate $\mathbf{x}_{i}$, timestamp with microsecond-level resolution $\tau_{i}$ and polarity $p_{i}\in \{-1,1\}$ of the brightness change. 
We refer to~\cite{Gallego20pami} for more information about the working principles of event cameras.
Given the reference patch $\mathbf{P}_0$, our network predicts the relative feature displacement $\mathbf{\Delta \hat{f}_{j}}$ during $t_{j-1}$ and $t_j$ using the corresponding event stream $E_j$ in the local neighborhood of the feature location at the previous timestep $t_{j-1}$.
The events inside the local window are converted to a dense event representation $\mathbf{P}_j$, specifically a maximal timestamp version of SBT~\cite{sbt} where each pixel is assigned the timestamp of the most recent event.
Once our network has localized the reference patch $\mathbf{P}_0$ inside the current event patch $\mathbf{P}_j$, the feature track is updated, and a new event patch $\mathbf{P}_{j+1}$ is extracted at the newly predicted feature location while keeping the reference patch $\mathbf{P}_0$.
This procedure can then be iteratively repeated while accumulating the relative displacements to construct one continuous feature track.
The overview of our method and our novel frame attention module are visualized in Fig.~\ref{fig:method}
In addition to temporal feature tracking, the localization of the grayscale patch inside the event stream can be applied to the task of sparse disparity estimation.
Instead of estimating the temporal displacement, the same architecture can predict the spatial displacement between a 3D point observed by an event and a standard camera in a side-by-side setting.
This adaptation to the disparity estimation task allows us to infer the depth of tracked features, which is vital for VO/SLAM. 
Given the flexibility of our architecture, only small modifications are needed for this additional task.
\reviewnew{
The shared architecture for both feature tracking and disparity estimation is introduced in Sec.~\ref{sec:architecture}.
Sec.~\ref{sec:feature_network} explains how the feature network processes each track independently, while Sec.~\ref{sec:frame_attention} details the frame attention module for incorporating global image information. 
For the specific network details, we refer to the supplementary.
Our feature tracking supervision with synthetic data and pose supervision strategy are covered in Sec.~\ref{sec:supervision}, and the adaptation of the architecture for disparity estimation is presented in Sec.~\ref{sec:disparity_estimation}.
}

\reviewnew{
\subsection{Architecture}
\label{sec:architecture}
\subsubsection{Feature Network}
\label{sec:feature_network}
}
To localize the template patch $\mathbf{P}_0$ inside the current event patch $\mathbf{P}_j$, the feature network first encodes both patches using separate encoders based on Feature Pyramid Networks~\cite{fpn}.
The resulting outputs are per-pixel feature maps for both patches that contain contextual information while keeping the spatial information.
To explicitly compute the similarity measure between each pixel in the event patch and the template patch, we construct a correlation map $C_j$ based on the bottleneck feature vector $R_0$ of the template patch encoder and the feature map of the event patch, as visualized in Fig.~\ref{fig:method}.
Together with the correlation map $C_j$, both feature maps are then given as input to a second feature encoder in order to refine the correlation map.
This feature encoder consists of standard convolutions, and one ConvLSTM block~\cite{convlstm} with a temporal cell state $F_j$.
The temporal information is crucial to predicting consistent feature tracks over time.
Moreover, it enables the integration of the motion information provided by the events.
The output of the feature network is a single feature vector with spatial dimension 1$\times$1.
Up to now, each feature has been processed independently from each other.

\reviewnew{
\subsubsection{Frame Attention Module}
\label{sec:frame_attention}
}
To share information between features in the same image, we introduce a novel \textit{frame attention module}, which is visualized in Fig.~\ref{fig:method}.
Since points on a rigid body exhibit correlated motion in the image plane, there is a substantial benefit in sharing information between features across the image.
To achieve this, our frame attention module takes the feature vectors of all patches at the current timestep $t_j$ as input and computes the final displacement for each patch based on a self-attention weighted fusion of all feature vectors.
Specifically, we maintain a state $S$ for each feature across time in order to leverage the displacement prediction of the previous timesteps in the attention fusion.
The temporal information should facilitate the information-sharing of features with similar motion in the past.
This way, it is possible to maintain vulnerable feature tracks in challenging situations by adaptively conditioning them on similar feature tracks.
Each input feature vector is first individually fused with the current state $S_{j-1}$ using two linear layers with Leaky ReLU activations (MLP).
All of the resulting fused features in an image are then used as key, query, and value pairs for a multi-head attention layer (MHA)~\cite{Vaswani17neurips}, which performs self-attention over each feature in an image.
To facilitate the training, we introduce a skip connection around the multi-head attention for each feature, which is adaptively weighted during the training by a Layerscale layer~\cite{Touvron2021ICCV} (LS).
The resulting feature vectors are then used in a simple gating layer to compute the updated state $S_j$ based on the previous state $S_{j-1}$ (GL), see Eq.~\ref{eq:mha}. 
\begin{align}
    Z_j^k  &= MLP([F_j, S_{j-1}])  \\
    \Tilde{Z}_j^k  &= MHA(Z_j^k)  \\
    S_j &= GL([S_{j-1}, LS(\Tilde{Z}_j^k)]) \label{eq:mha}
\end{align}
Finally, the updated state $S_j$ is then processed by one linear layer to predict the final displacement $\mathbf{\Delta \hat{f}_j}$.
\reviewnew{
\subsection{Feature Tracking Supervision}
\label{sec:supervision}
}
In general, the supervision of trackers, extractors, or even flow networks is still an open research field since datasets containing pixel-wise correspondences as ground truth are rare.
To make matters worse, there exist even fewer event-based datasets containing accurate pixel correspondences.
To overcome this limitation, we train our network in the first step on synthetic data from the Multiflow dataset~\cite{multiflow}, which contains frames, synthetically generated events, and ground truth pixel flow.
However, since the noise is not modeled, synthetic events differ significantly from events recorded by a real event camera.
Thus, in the second step, we fine-tune our network using our novel pose supervision loss to close the gap between synthetic and real events.

\textbf{Synthetic Supervision}
Synthetic data has the benefit that it provides ground truth feature tracks.
Thus, a loss based on the L1 distance can be directly applied for each prediction step $j$ between the predicted and ground truth relative displacement, see Fig. 2 in the supplementary. 
It is possible that the predicted feature tracks diverge beyond the template patch such that the next feature location is not in the current search.
Thus, if the difference between predicted and ground truth displacement $||\mathbf{\Delta \hat{f}_{j}} - \mathbf{\Delta f_{j}}||_1$ exceeds the patch radius $r$, we do not add the L1 distances to the final loss to avoid introducing noise in supervision.
Our truncated loss $\mathcal{L}_{rp}$ is formulated as follows.
\begin{gather}
    err_{j} = % 
  \begin{cases}
    ||\mathbf{\Delta \hat{f}_{j}} - \mathbf{\Delta f_{j}}||_1 \quad \quad & ||\mathbf{\Delta \hat{f}_{j}} - \mathbf{\Delta f_{j}}||_1 < r\\
    0 &\text{else}
  \end{cases}  \label{loss_reproj_start}\\
    \mathcal{L}_{rp} = \frac{\sum_{j} \mathbbm{1}_{(err_{j} \neq 0)} err_{j}}{\sum_{j} \mathbbm{1}_{(err_{j} \neq 0)}} \label{loss_reproj_end} 
\end{gather}

To reduce the gap between synthetic and real data, we apply on-the-fly augmentation during training, which significantly increases the motion distribution. 
To teach the network geometrically robust representations, affine transformations $W$ are applied to the current event patch $\mathbf{P_j}$ to obtain an augmented Patch $\mathbf{P^{aug}_j}$ at each prediction step, as formulated in Eq.~\ref{eq:track_augm}.
The augmentation parameters for rotation, translation, and scale $\boldsymbol{\theta} = (\theta_{r}, \theta_{t}, \theta_{s})$ are randomly sampled from a uniform distribution at each prediction step during training. 
Our tracker $T$ then predicts a relative displacement $\mathbf{\Delta \hat{f}^{aug}_{j-1}}$ given the augmented patch $\mathbf{P^{aug}_j}$ and original template patch $\mathbf{P_0}$.
The loss is then computed between the predicted displacement $\mathbf{\Delta \hat{f}^{aug}_{j-1}}$ and the augmented ground truth $\mathbf{\Delta f^{aug}_{j-1}}$, which is obtained by applying the same affine transformation $W$.
\begin{align}
    \mathbf{P^{aug}_j} &= W(\mathbf{P_j}, \boldsymbol{\theta}) \label{eq:track_augm}\\
    \mathbf{\Delta \hat{f}^{aug}_{j-1}} &= T(\mathbf{P_0},\mathbf{P^{aug}_j})  \\
    \mathbf{\Delta \hat{f}_{j-1}} &= W^{-1}(\mathbf{\Delta \hat{f}^{aug}_{j-1}}, \boldsymbol{\theta}) \label{eq:inv_augm}
\end{align}
The corrected displacement $\mathbf{\Delta \hat{f}_{j-1}}$ is then accumulated to extract the next event patch $\mathbf{P_{j+1}}$.
Our augmentation strategy introduces dynamic trajectories and changes in patch appearance that improve performance on real data.

\begin{figure}[t!]
\centering
\includegraphics[width=0.40\textwidth]{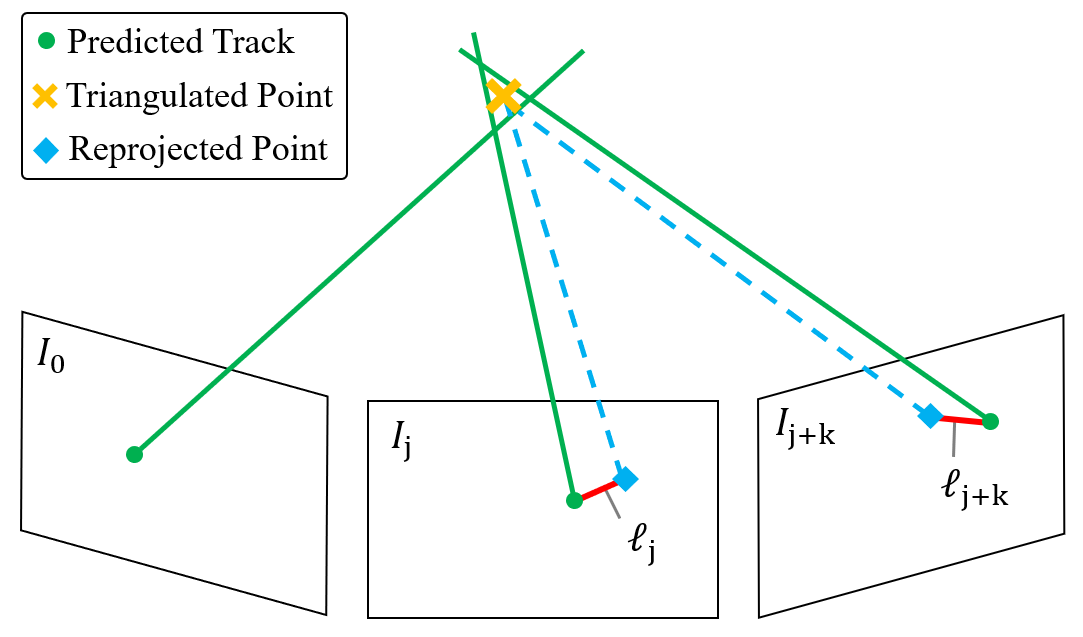}
\vspace{-5pt}
\caption{
To adapt our tracker to real event data, our self-supervised loss computes a triangulated point based on the predicted track, and the camera poses.
The 3D point is then reprojected to each camera plane, and the L1-distance $\ell_j$ between reprojected and predicted point is used as a supervision signal.
}
\label{fig:supervision_triangulation}
\end{figure}

\textbf{Pose Supervision}
To adapt the network to real events, we introduce a novel pose supervision loss solely based on ground truth poses of a calibrated camera.
The ground truth poses can be obtained for sparse timesteps $t_j$ using structure-from-motion algorithms, e.g., COLMAP~\cite{colmap_sfm}, possibly combined with inertial measurements~\cite{cioffi2022continuous} or by an external motion capture system to achieve high robustness in challenging scenarios.
Since our supervision strategy relies on the triangulation of 3D points based on poses, it can only be applied in static scenes.
In the first step of the fine-tuning, our network predicts multiple feature tracks for one sequence. 
For each predicted track $i$, we compute the corresponding 3D point $\mathbf{X}_i$ using the direct linear transform~\cite{abdel2015direct}, which is explained in the supplementary.
Once the 3D position of $\mathbf{X}_i$ is computed, we can find the reprojected pixel point $\mathbf{\Tilde{x}}_j$ for each timestep $t_j$.
The final pose supervision loss is then constructed based on the predicted feature $\mathbf{\hat{x}_{j}}$ and the reprojected feature $\mathbf{\Tilde{x}}_j$ for each available camera pose at timestep $t_j$, as visualized in Fig.~\ref{fig:supervision_triangulation}.
As in the supervised setting of Eq.~\ref{loss_reproj_end}, we use a truncated loss, which excludes the loss contribution if the reprojected feature is outside of the event patch.

\subsection{Sparse Disparity Estimation}
\label{sec:disparity_estimation}
\begin{figure}[ht!]
\centering
\includegraphics[width=0.48\textwidth]{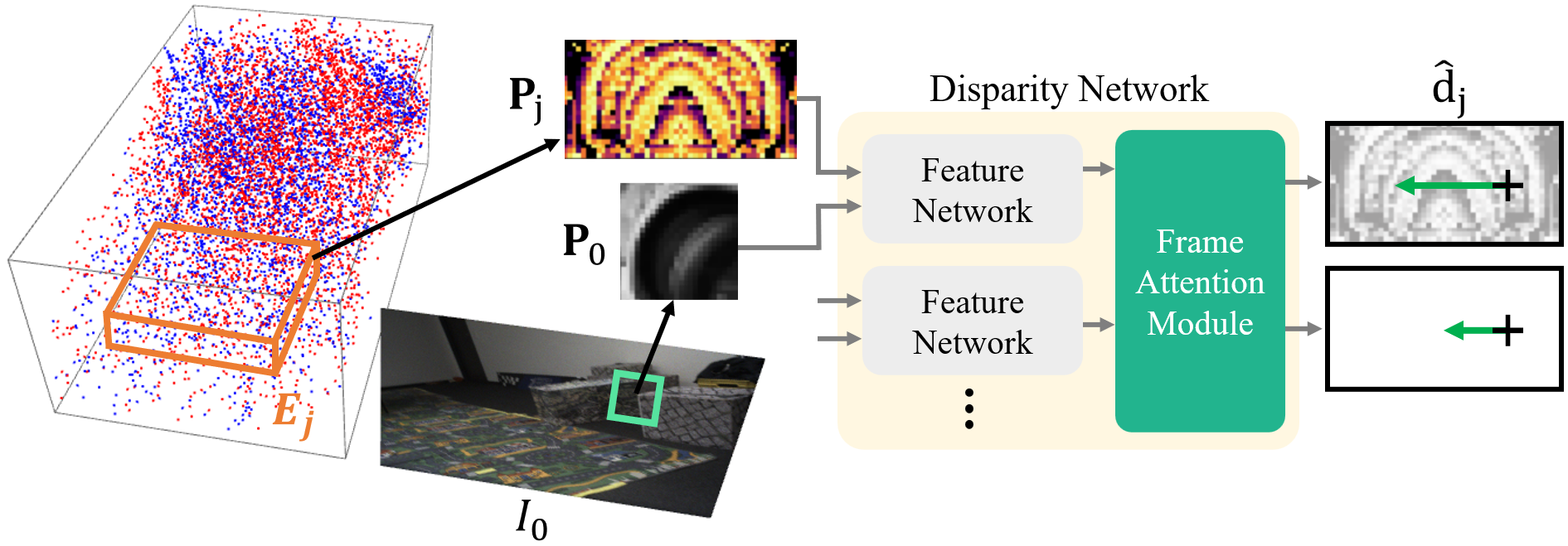}
\caption{
For the disparity estimation task, our proposed network takes as input a reference patch $\mathbf{P_0}$ in a grayscale image $I_0$ and a rectangular event patch $\mathbf{P}_j$ constructed from an event stream $E_j$ at timestep $t_0$.
Similar to the feature tracking task, our novel frame attention module enables the information sharing across different features to compute the final disparity $\mathbf{\hat{d}_{j}}$.
}
\label{fig:disp_method}
\end{figure} 

Beyond tracking features over time, our proposed approach to locate a grayscale image patch within an event voxel grid can be employed for estimating the sparse disparities between features captured by an event and a standard camera in a rectified side-by-side configuration. 
Similar to feature tracking, we predict the disparity values for a set of sparse image points tracked over time, i.e. feature tracks. 
The adaptation of our proposed feature tracking method to disparity estimation offers a fresh perspective on the task.
For instance, although our disparity method does not have access to full frames, our proposed frame attention module combines global frame information from the sparse patches while accumulating temporal information along feature tracks.
Moreover, the disparities of image points within any given frame share information because of the natural structure of the 3D world.
Therefore, the ability of the frame attention module to share information across image features within the same frame can improve the accuracy and robustness of our disparity estimation.

Our feature tracking methodology seamlessly extends to disparity estimation with minimal adjustments to the inputs and network layers. 
Operating within a rectified dual-camera setup, the search space in the event stream is constrained to a single spatial dimension, transforming the quadratic search patch into a rectangle aligned along the epipolar lines while keeping a quadratic grayscale patch as a template.
Different from the dynamic event patch $\mathbf{P}_j$ used in feature tracking, we maintain a fixed position for the rectangular event patch, ensuring coverage across a predefined disparity range.
The fixed position removes the dependency on the previous disparity estimate, leading to accelerated data loading and enhanced training stability.
To account for the larger search space, we increase the contextual information by enlarging the grayscale patch size from 31 to 63 and using a $61\times122$ event patch.
A convolution layer with stride two in a single dimension is applied to both the rectangular correlation $C_j$ and event feature map, aligning them with the spatial dimension of the grayscale feature map.
Additionally, to introduce positional information to the sparse event patch, we concatenate the feature position in the form of a two-channel coordinate grid to the grayscale template patch.
Fig.~\ref{fig:disp_method} illustrates the adapted setup for sparse disparity estimation.
\newcolumntype{C}[1]{>{\centering}m{#1}}

\section{Feature Tracking Experiments}
\label{sec:experiments_feature}

\textbf{Datasets} 
We compare our proposed data-driven tracker on the commonly used Event Camera dataset~\cite{Mueggler17ijrr} (EC), which includes APS frames (\unit[24]{Hz}) and events with a resolution of 240$\times$180, recorded using a DAVIS240C camera~\cite{Brandli14ssc}.
Additionally, the dataset provides ground truth camera poses at a rate of \unit[200]{Hz} from an external motion capture system.
Moreover, to evaluate the tracking performance with a newer sensor setup, we test our method on the newly published Event-aided Direct Sparse Odometry dataset~\cite{Hidalgo2022cvpr} (EDS).
Compared to EC, the EDS dataset contains higher resolution frames and events (640$\times$480 pixels) captured with a beam splitter setup.
Similar to the EC dataset, it includes ground truth poses at a rate of \unit[150]{Hz} from an external motion capture system.
Most scenes in both datasets are static since the primary purpose of EDS and EC is the evaluation of camera pose estimation.
For the specific finetuning and testing sequence selection, we refer to the supplementary.
\textbf{Evaluation}
To evaluate the different feature trackers, we first extract features for each sequence with a Harris Corner detector~\cite{Harris88}.
Based on the initial feature set, each tested tracker predicts the feature displacements according to its specific update rate.
Unfortunately, no ground truth feature tracks are available for EDS and EC.
To evaluate the event-based feature trackers without ground truth, previous works used tracks predicted by the frame-based KLT tracker as ground truth.
Instead, to increase the accuracy of KLT tracks, we use an evaluation scheme based on our proposed pose supervision method.
Specifically, the ground truth tracks are obtained by triangulating KLT tracks using ground truth poses and reprojecting them afterward to each of the selected target frames.
The triangulation of KLT tracks has the benefit that minor tracking errors of KLT are filtered out, leading to geometrically consistent ground truth tracks.
To verify the proposed evaluation, we conducted an experiment in simulation in which ground truth feature tracks are available.
In this simulated setup, we computed the Pearson correlation between the KLT reprojected error and the ground truth feature tracks, which was $0.716$.
This indicates a significant correlation between our proposed evaluation technique and ground truth feature tracks verifying the effectiveness of our evaluation technique.
Since each tested tracker has its update rate, we linearly interpolated all feature tracks to the ground truth pose timesteps in order to compute the evaluation metric.
Furthermore, to effectively test the event-based tracking abilities of the methods, we do not update the feature templates during evaluation.
In addition, we deactivate any terminal criterion and report the time until the feature exceeds a certain distance to the ground truth, known as the feature age.
Instead of choosing one error threshold as done in previous work~\cite{Alzugaray2020HASTE:Events}, we evaluate the resulting tracks for multiple error thresholds in a range from 1 to 31 pixels with a step size of 1 pixel.
Thus, we do not report the endpoint error since we test each trajectory with different error thresholds, which effectively incorporates the distance error into the feature age.
As a first performance metric, we compute the tracked feature age normalized by the ground truth track duration in order to account for different trajectory lengths.
However, since some feature tracks are lost immediately in the beginning, we report the feature age of stable tracks, i.e., we discard feature tracks lost during the early phase of the sequence for the feature age computations.
The second error metric accounts for the lost tracks by taking the ratio of stable tracks and ground truth tracks.
This ratio is then multiplied by the feature age, which gives us the expected feature age as the second performance metric.
This metric combines the quality and the number of feature tracks tracked by a method.
For more information about the two performance  metrics, we refer to the supplementary.
\textbf{Training Schedule}
As mentioned in Sec.~\ref{sec:method}, we first train our models supervised on the Multiflow~\cite{multiflow} dataset on 30000 feature tracks in a continual learning fashion with a learning rate of $1\times 10^{-4}$ using the ADAM optimizer~\cite{Kingma15iclr} to gradually adapt the network recurrence to longer trajectory lengths.
Starting initially from 4 unroll steps, we progressively increase the number of unroll steps to 16 and then 24 after 80000 and 120000 training steps, respectively.
After training on Multiflow, we finetune our model using our novel supervision method for 700 optimization steps with a reduced learning rate of $1\times 10^{-6}$ on specific training sequences of both datasets, which are not used for evaluation. 
\reviewnew{
\subsection{Aligned Events and Frames Tracker}
\label{sec:combination_events_frames}
}
\begin{table*}[ht!]
\footnotesize
\centering
\caption{
The performance of the evaluated trackers on the EDS and EC dataset are reported in terms of "Feature Age (FA)" of the stable tracks and the "Expected FA", which is the multiplication of the feature age by the ratio of the number of stable tracks over the number of initial features.
}
\begin{tabular}{m{3.1cm}C{0.9cm}C{2.8cm}C{2.7cm}C{2.8cm}>{\centering\arraybackslash}m{2.7cm}}
   &  & \multicolumn{2}{c}{EDS} & \multicolumn{2}{c}{EC} \\
\cmidrule(lr){3-4} \cmidrule(lr){5-6}
Method & Frames & Feature Age (FA) $\uparrow$ & Expected FA $\uparrow$ & Feature Age (FA) $\uparrow$ & Expected FA $\uparrow$  \\
\hline
ICP~\cite{Kueng16iros}                     & no  & 0.060 & 0.040 & 0.256 & 0.245 \\
EM-ICP~\cite{Zhu17icra}                    & no  & 0.161 & 0.120 & 0.337 & 0.334 \\
HASTE~\cite{Alzugaray2020HASTE:Events}     & no  & 0.096 & 0.063 & 0.442 & 0.427 \\
EKLT~\cite{eklt}                           & yes & 0.325    & 0.205    & \underline{0.811} & 0.775 \\
\textbf{Ours} (zero-shot)                  & yes & 0.549    & 0.451    & 0.795             & \underline{0.787} \\
\textbf{Ours} (fine-tuned)                 & yes & 0.576    & 0.472    & \textbf{0.825}    & \textbf{0.818} \\
\hline
\textbf{Ours} E2VID (transfer)             & no & 0.524 & 0.435 & 0.709 & 0.695 \\
\textbf{Ours} E2VID (zero-shot)            & no & \underline{0.579} & \underline{0.482} & 0.793 & 0.781 \\
\textbf{Ours} E2VID (fine-tuned)           & no & \textbf{0.589} & \textbf{0.495} & 0.794 & 0.786
\end{tabular}
\label{tab:baseline_results}
\end{table*}

\textbf{Baselines}
We compare our method against the current state-of-the-art method EKLT~\cite{eklt}, which extracts a template patch from a grayscale image for each feature and tracks the feature with events, similar to our tracker.
As another tracker relying on grayscale template patches, we also run the ICP~\cite{Kueng16iros} tracker used for event-based visual odometry.
In addition, we evaluate against the pure event-based trackers HASTE~\cite{Alzugaray2020HASTE:Events} and EM-ICP~\cite{Zhu17icra}.
For EKLT, HASTE, and EM-ICP, we adopted the publicly available code to run the experiments.
The implementation of ICP was taken from a related work~\cite{Dong21acm}.
The hyper-parameters of all methods were tuned for the specific datasets, which required multiple hours to achieve optimal performance.

\begin{figure}[tb!]
\centering
\includegraphics[width=0.47\textwidth]{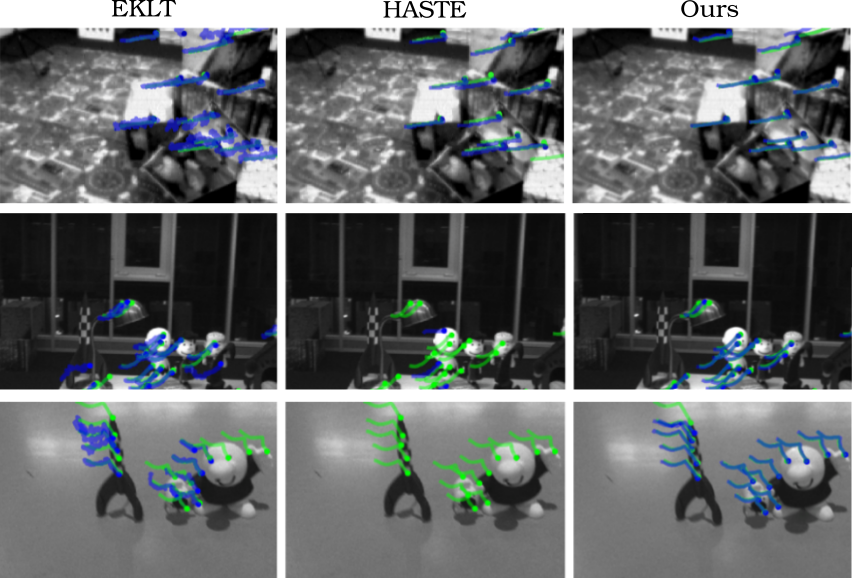}
\caption{
Qualitative tracking predictions (blue) and ground truth tracks (green) for the EC dataset (top) and EDS dataset (middle / bottom).
Our method predicts more accurate tracks for a higher number of initial features.
}
\label{fig:qualitative_ec_eds}
\end{figure}

\textbf{EC Results}
On the commonly used event-based tracking benchmark, EC, our proposed data-driven method with grayscale template patch, i.e., Ours (zero-shot) and Ours (fine-tuned), outperforms the other baselines in terms of non-zero feature age and expected feature age, see Tab.~\ref{tab:baseline_results}.
The second best approach is EKLT, which tracked the features for a duration similar to our proposed method as represented by the non-zero feature age metric in Tab.~\ref{tab:baseline_results}.
However, our method was able to track more features from the initial feature set as reported by the expected feature age.
The higher ratio of successfully tracked features and the longer feature age makes our method better suited for downstream tasks such as pose estimation~\cite{colmap_sfm}.
The top row of Fig.~\ref{fig:qualitative_ec_eds} shows that our method produces a higher number of smooth feature tracks compared to the closest baselines EKLT and HASTE.
As expected, a performance gap exists between pure event-based methods (HASTE, EM-ICP) and methods using grayscale images as templates (Ours, EKLT).
This confirms the benefit of leveraging grayscale images to extract template patches, which are subsequently tracked by events.

\textbf{EDS Results}
Similar to the performance on the EC dataset, our proposed method with grayscale template patch, i.e., Ours (zero-shot) and Ours (fine-tuned), outperforms all of the existing trackers on the EDS dataset with an even larger margin in terms of both non-zero feature age and expected feature age as reported in Tab.~\ref{tab:baseline_results}.
The significant performance boost confirms the capability of our data-driven methods to deal with high-resolution data in various 3D scenes with different lighting conditions and noise patterns.
Since a beam splitter setup was used to record the data for the EDS dataset, there are misalignment artifacts between events and images, as well as low-light noise in the events due to the reduction of the incoming light.
Additionally, the EDS includes faster camera motions leading to an overall lower tracking performance of all methods compared to the EC dataset.
Nevertheless, our learned method is able to deal with these different noise sources and still predict smooth feature tracks for a large number of features, as shown in the middle and bottom row of Fig.~\ref{fig:qualitative_ec_eds}.
For more qualitative examples, we refer to the supplementary.
In addition to the performance gain, our method does not require hours of manual fine-tuning to transfer the tracker from small resolution to high-resolution event cameras with different contrast threshold settings.
\reviewnew{
Finally, our event-based tracker can provide robust feature tracks during periods of high-speed motion in which the frames suffer from motion blur, as illustrated in Fig.~\ref{fig:eye_catcher}.
This high-rotational motion sequence was recorded by us with a beam splitter setup.
}

\begin{figure}[tb!]
\centering
\includegraphics[width=0.47\textwidth]{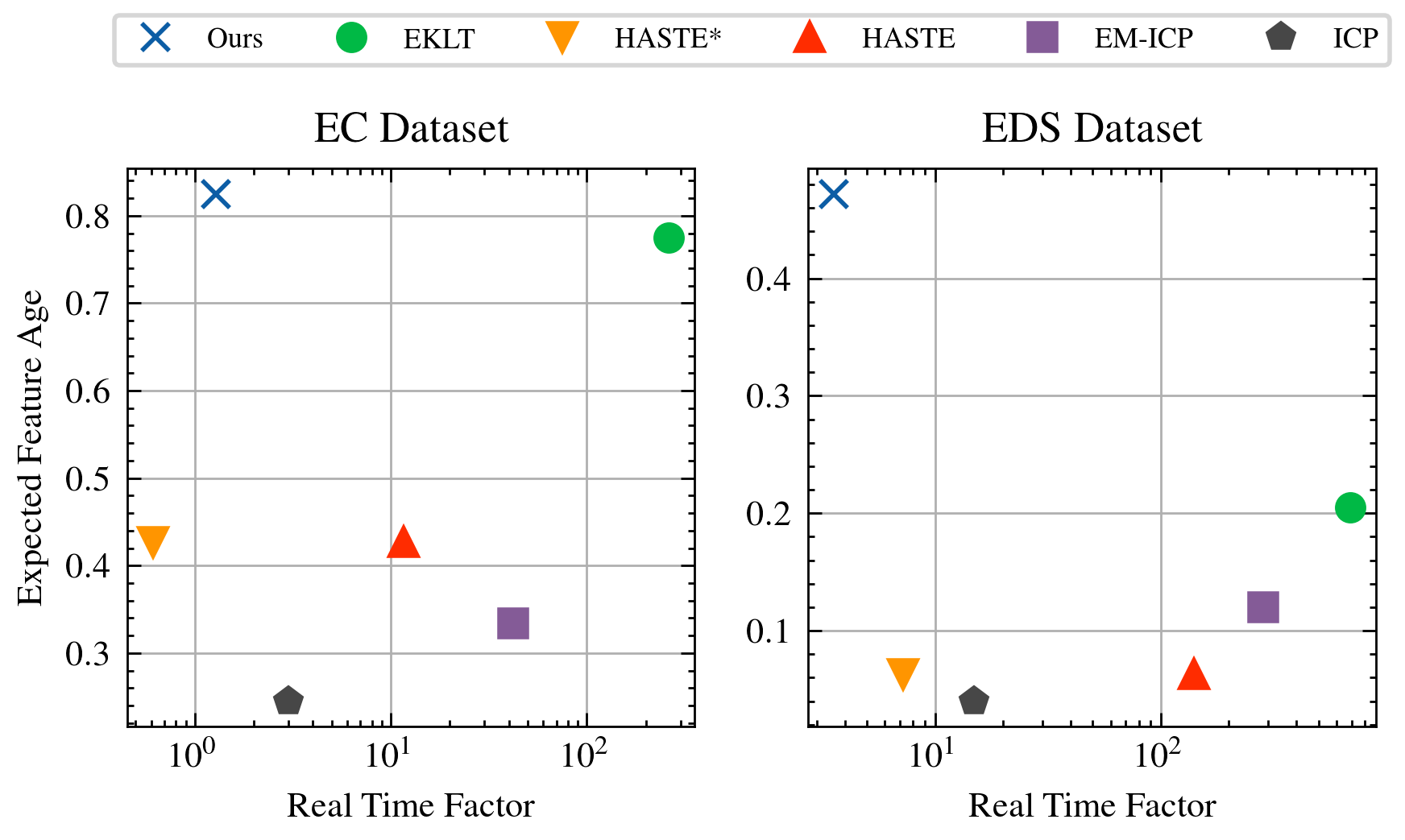}
\vspace*{-8pt}
\caption{
The two plots show the tracking performance in terms of expected feature age in relation to the real-time factor, which is the ratio of compute time over track time.
Thus, the top left corner represents the goal.
Additionally to the existing implementation of HASTE, we also report the ideal HASTE*, which assumes perfect parallelization for processing all feature tracks.
}
\label{fig:runtime}
\end{figure}

\textbf{Runtime Comparison}
To employ a feature tracker in real-world applications, it is crucial to provide feature displacement updates with low latency.
Therefore, we report the runtime of the different evaluated methods in terms of the real time factor, i.e., compute time divided by the time of the received data, versus tracking performance in Fig.~\ref{fig:runtime}.
It should be noted that most of the evaluated trackers were not implemented for run time efficiency and thus are coded in different programming languages, which makes a fair comparison hard.
Moreover, we tuned all the methods with a focus on the tracking performance, which explains the high runtime of EKLT since we significantly increased the number of optimization iterations.
Nevertheless, the runtime comparison of the different methods still provides a rough picture of the inference speed of each method. 
In the case of HASTE, we additionally report the runtime for an ideal HASTE implementation, named HASTE* in Fig.~\ref{fig:runtime}.
The ideal HASTE* assumes perfect parallelization of the current code framework of HASTE, which tracks each feature sequentially.
Even without optimizing the code for deployment, our method achieves close to real-time performance on EC and is the fastest method on EDS while having a significantly higher tracking performance.
On EDS, our method takes 17ms to process, on average, 19.7 patches in parallel, while it takes 13ms for 14.2 patches on EC using an Nvidia Quadro RTX 8000 GPU.
The fast inference of our method can be explained by the batch-wise processing and the highly parallelized framework for deep learning architectures.
This shows the potential of our method for real-world applications

\reviewnew{
\textbf{Combination with frame-based KLT}
In a step to combine the contextual information of grayscale images and the high-latency information from events, we extended our event-based tracker using the popular KLT tracker for frames. 
Specifically, we use our event tracker to track features during the blind time between two frames and use the displacement prediction of our tracker as an initial guess for the KLT tracker once a new frame arrives.
This has the benefit of effectively mitigating the negative effects of large baselines between two frames caused by high-speed motion.
The improved tracking performance is reported in Fig.~\ref{fig:frameskip} and further discussed in the supplementary material.
\begin{figure}[tb!]
\centering
\includegraphics[trim={12.5cm 0 0 0},clip,width=0.28\textwidth]{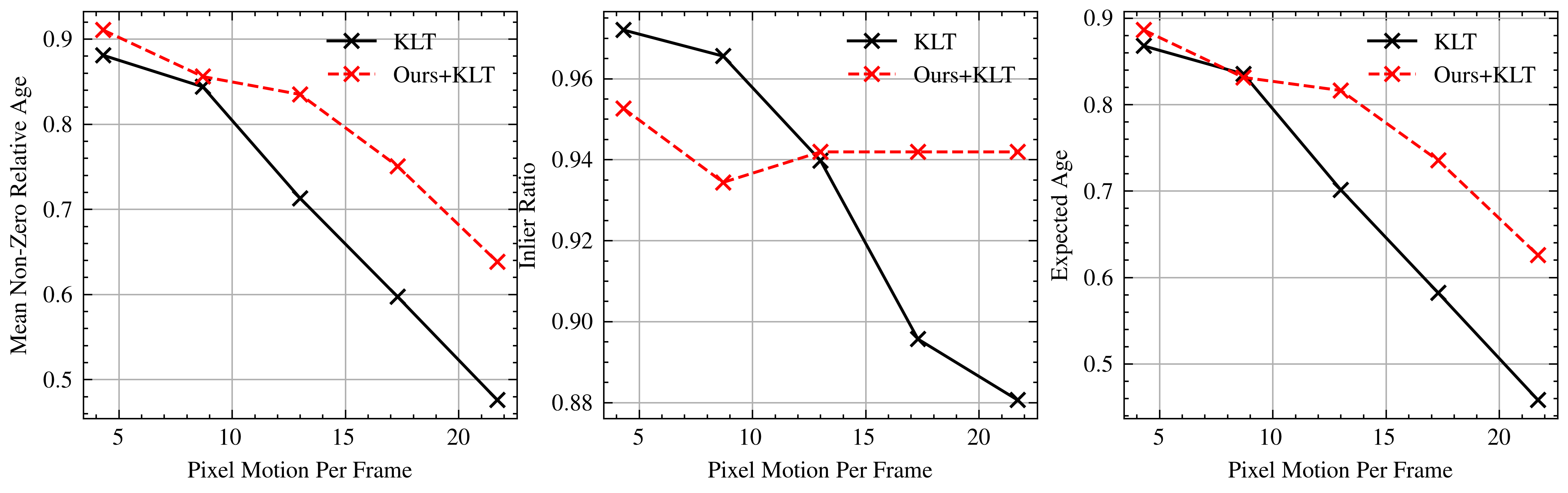}
\vspace*{-8pt}
\caption{
The tracking performance of KLT and our tracker combined with KLT (Ours+KLT) in relation to pixel motion per frame.
In combination, our event-based tracker can successfully help KLT predict larger displacement while KLT can refine the predictions of our tracker.
}

\label{fig:frameskip}
\end{figure} 
}

\rreviewnew{
\textbf{Ablations}
We present ablation studies in the supplementary to demonstrate the benefits of the training augmentation, the frame attention module, and other network design choices.
In the supplementary, we also report the results for the input representation and augmentation parameters ablations.
}

\reviewnew{
\subsection{Tracker Without Frames}
}
To eliminate the requirement for two spatially aligned cameras, we tested our approach by replacing grayscale images with event-to-frame reconstructions using E2VID~\cite{Rebecq19pami}.
In Tab.~\ref{tab:baseline_results}, we report the performance of our tracker under three training settings, with evaluations always conducted on E2VID reconstructions: (i) trained with grayscale frames, denoted as Ours E2VID (transfer); (ii) trained with E2VID reconstructions, denoted as Ours E2VID (zero-shot); and (iii) trained and fine-tuned with E2VID reconstructions, employing our pose supervision technique, denoted as Ours E2VID (fine-tuned). 

\textbf{EC Results}
Our event-only tracker outperforms other event-based trackers on the EC dataset, although falling short of the performance of EKLT, which uses both events and frames.
Similar to our method with grayscale frames, our proposed pose supervision technique enhances the performance of our pure event-based tracker. 
As expected, training on E2VID reconstructions leads to better performance compared to training on grayscale images and subsequently transferring to E2VID images during inference.
\textbf{EDS Results}
On the EDS dataset, all of our trackers using E2VID reconstructions significantly outperform the baselines. 
Interestingly, our trackers trained with E2VID reconstructions exceed the performance of our model leveraging grayscale frames.
This discrepancy can be attributed to slight misalignments between events and frames caused by the beam splitter setup employed in the EDS dataset.
This explanation is supported by results from the EC dataset, where the DAVIS sensor ensures perfect alignment. 

\reviewnew{
\textbf{Runtime}
Using E2VID requires a constant runtime of 153.33ms for each reconstructed frame. 
However, the E2VID reconstructions can be generated in parallel with the event tracker updates. 
In this setup, the event tracker continues to use the previous E2VID template patch while the new E2VID frame is generated, replacing it once the reconstruction is complete. 
Therefore, assuming perfect parallelization, the latency and runtime of the event tracker will not be influenced by the E2VID reconstruction. 
}

\section{Disparity Estimation Experiments}
\label{sec:experiments_disparity}
\newcommand{\cw}{1.1cm}
\newcommand{\cd}{0.75cm}

\begin{table}[ht!]
\footnotesize
\centering
\caption{
The disparity metrics on M3ED and the runtime that is required to process the tracks in one frame.
}
\scalebox{0.92}{
\begin{tabular}{m{2.2cm}C{0.9cm}C{\cw}C{\cd}C{\cd}>{\centering\arraybackslash}m{1.0cm}}
     & \multicolumn{4}{c}{Disparity Metrics} & Runtime\\
\cmidrule(lr){2-5} \cmidrule(lr){6-6}
Method  & MAE $\downarrow$ & RMSE $\downarrow$ & 1PE $\downarrow$ & 4PE $\downarrow$ & [ms] $\downarrow$  \\
\hline
FEStereo~\cite{chen24wacv}           & 0.962          & 3.016          & \textbf{0.145} & 0.023          &  25.2 \\
CREStereo~\cite{li2022practical}     & 0.920          & 1.829          & 0.247          & 0.033          & 354.8 \\
\textbf{Ours}                        & \textbf{0.813} & \textbf{1.209} & 0.269          & \textbf{0.011} &   \textbf{8.7} \\
\end{tabular}}
\label{tab:disparity_results}
\end{table}

\begin{figure}[tb!]
\centering
\includegraphics[width=0.47\textwidth]{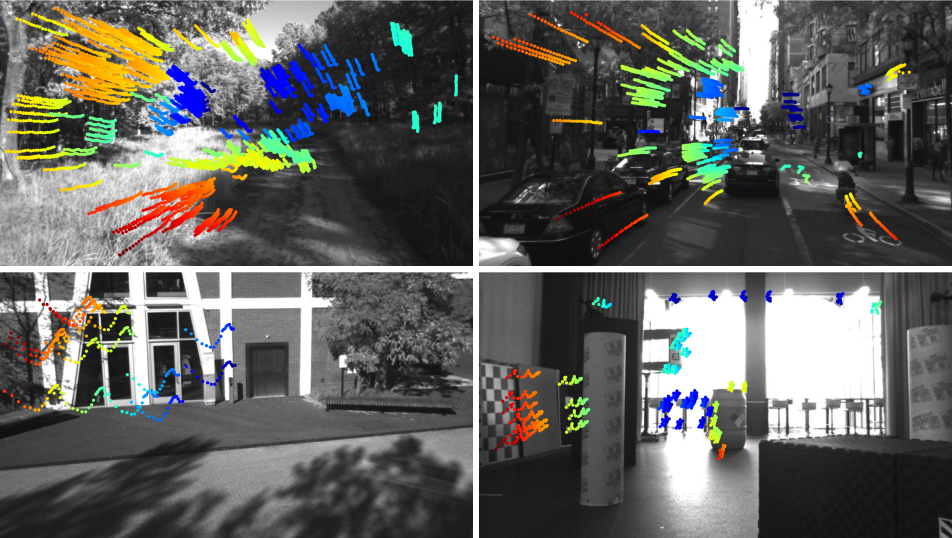}
\caption{
The tracks on the M3ED dataset are colorized based on their corresponding disparity values, ranging from blue to red to represent small to large disparity values.
}
\label{fig:disparity_tracks}
\end{figure}

\textbf{Datasets} 
To train our disparity network, we leverage the M3ED dataset~\cite{chaney23cvpr} featuring real-world scenes recorded with a stereo event and frame camera pair mounted on a quadrotor, a car, and a quadruped robot.
The dataset comprises diverse environments such as forest, urban, and indoor settings, providing a wide variety of scenes for training and testing.
For our approach, the left event and the left standard camera are used for a vertically rectified event and standard camera setup.
Ground truth feature tracks are generated using a Harris detector for feature detection, followed by tracking with a KLT tracker, which is combined with the depth and ground truth poses provided along the dataset.
Specifically, we perform forward tracking and subsequent backward tracking through the sequence using the KLT tracker on the frames to increase the duration of the feature tracks.
For each tracking step, the 3D positions of the feature tracks are updated whenever depth information is available. 
The reprojected 3D points serve as initial estimates for the KLT tracker and as termination criteria. Tracks are terminated when the error between tracked and reprojected points exceeds 3 pixels.
This approach ensures the generation of high-quality feature tracks complemented by accurate depth information for each track, see Fig.~\ref{fig:disparity_tracks}.
During training, we randomly sample twelve tracks with a length of 20 frame timesteps. 
To minimize overlap between sampled training batches, a margin of 10 timesteps is introduced between selected starting frames.

\textbf{Evaluation} 
For the test dataset, we employ the same technique used to construct the training dataset with the modification of selecting 16 feature tracks per starting frame and disregarding any additional tracks.
This adjustment ensures a more balanced test set since frames with a high number of tracks will contribute only 16 feature tracks. 
This results in a total of 1368 starting frames, each containing 16 feature tracks spanning a duration of 20 frames. 
The specific sequences used for training, validation, and test splits are listed in the supplementary material. 
In our evaluation, we compare the methods using common disparity metrics, including mean absolute error (MAE), root mean square error (RMSE), and the ratios of disparity errors exceeding one and four pixels (1PE / 4PE).
\textbf{Training} 
Similarly to the training of the learned feature tracker, we train the disparity network using the ADAM optimizer~\cite{Kingma15iclr} with a learning rate of $1\times 10^{-4}$ and use the L1 loss between the predicted and ground truth disparity.
Unlike the feature tracker, the search patch for the disparity network is not dependent on previous network predictions. 
As a result, there is no requirement for continual learning with increasing sequence length, allowing us to begin directly with 20 timesteps.
We adopt a batch size of four, resulting in the update of 48 tracks per batch, considering the 12 tracks per frame.
\subsection{Benchmark Results}
\textbf{Baselines} 
We evaluate our approach against existing methods that exploit the global context of both event representations and images.
This setup differs from ours, as the global context offers considerably more information compared to the cropped patches used in our method.
As our first baseline, we compare against the Frame-Event Stereo Matching module (FEStereo) from the DC-FEStereo approach~\cite{chen24wacv}, which is inspired by the design of an event stereo method~\cite{tulyakov18neurips}.
The other two modules of the DC-FEStereo method rely on ground truth poses and, therefore, are not applicable to our setting.
As our second baseline, we use the stereo image approach CREStereo~\cite{li2022practical} by converting the event streams first to frames using E2VID~\cite{Rebecq19pami}.
Both baselines are trained on the M3ED dataset with the provided sparse ground truth depth. 
We adopt the respective designed loss functions for the baselines and use default parameters for both approaches, except for the batch size of CREStereo, which we adjust to three to accommodate processing full-resolution images.
\textbf{Results} 
\rreviewnew{
Our sparse disparity network outperforms the dense baselines FEStereo and CREStereo in terms of MAE and RMSE, see Tab.~\ref{tab:disparity_results}. 
However, FEStereo and CREStereo have a lower ratio of disparity errors above 1 pixel, whereas our method only achieves better performance at error ratios above 4 pixels.
This indicates that both baselines predict more accurate disparities for a high portion of the tested tracks while suffering from large outlier errors, which also explains the high RMSE value.
}
The large outliers can be explained by the initialization of feature tracks with a Harris detector, which results in features mostly located at depth discontinuous between foreground and background.
Our method leverages the temporal context specific to the feature tracks, which makes the distinction between fore- and background easier and thus reduces the outliers.
A potential downside of using a stereo system with a frame camera is the decreasing frame quality in challenging scenarios. 
However, if the downstream task of pose estimation is considered, our event-based feature tracker can be used to track features while localizing in a map constructed with the sparse disparity module using frames captured under better conditions.
\textbf{Runtime Comparison}
The significant benefit of our method lies in the sparse patch processing, resulting in significantly faster runtimes compared to the dense baselines, as reported in Tab.~\ref{tab:disparity_results}.
Processing 16 tracks with our method requires only 8.7ms, compared to the 25.2ms required by the dense FEStereo method to process a full image.
\rreviewnew{
Our method is even 40 times faster compared to the E2VID+CREStereo approach, which requires running E2VID (153.33ms) and CREStereo (201.48ms) sequentially.
}
\reviewnew{
Since our network processes feature tracks mostly in parallel, there is only a small runtime increase when scaling up the number of tracks, as reported in the supplementary.
}

\rreviewnew{
\textbf{Stereo-VO}
To provide an initial evaluation of the effectiveness of our approach for pose estimation as a downstream task, we construct a basic stereo-VO pipeline.
Similarly to the disparity evaluation, we test on the M3ED dataset by leveraging the same features detected with the Harris detector.
First, the detected features are back-projected into 3D space by using either disparities predicted by our method or the ground truth disparity.
Next, the 3D map is used to localize the camera at the subsequent timesteps by either using the ground truth features tracks or the feature tracks computed with our proposed event-based feature tracker combined with frames, see Sec.~\ref{sec:combination_events_frames}.
Camera pose estimation was performed using the PnP algorithm~\cite{lepetit2009ep}, which aligns 3D points with their corresponding 2D image points.
To assess the impact of our disparity estimation method and our tracker, we conduct tests where either the disparity or the feature tracks were replaced with ground truth data.
Table~\ref{tab:sup_stereovo} reports the pose estimation performance in terms of Absolute Trajectory Error for translation (ATE Trans.) and rotation (ATE Rot.).
Using ground truth tracks with predicted disparities achieves the lowest translational error but the highest rotational error, both of which remain small.
Replacing predicted disparities with ground truth (Pred Tracks / GT Disp) yields a similar ATE Trans. of 2.5 cm while reducing rotational error.
The performance of the full stereo VO pipeline (Pred Tracks / Pred Disp) is within 8 mm of pipelines incorporating ground truth data, demonstrating the effectiveness of our approach for pose estimation.
}

\begin{table}[tb!]
\footnotesize
\caption{
Stereo VO performance using feature tracks from our tracker (Pred Tracks) or ground truth tracks (GT Tracks), combined with either predicted disparities (Pred Disp) or ground truth disparities (GT Disp).
}
\centering
\begin{tabular}{m{3.5cm}C{1.5cm}>{\centering\arraybackslash}m{1.5cm}}
Method  & ATE Trans. [m] $\downarrow$ & ATE Rot. [$^{\circ}$] $\downarrow$ \\
\hline
GT Tracks / Pred Disp       & 0.024 & 2.05 \\
Pred Tracks / GT Disp       & 0.025 & 1.61 \\
Pred Tracks / Pred Disp     & 0.033 & 1.74 \\
\end{tabular}
\label{tab:sup_stereovo}
\vspace*{-10pt}
\end{table}

\section{Conclusion}
\label{sec:Conclusion}
We presented the first data-driven feature tracker for event cameras, which leverages low-latency events to track features detected in a grayscale frame.
Our tracker is designed to operate with both a spatially aligned image and event camera setup, as well as a single event camera setup.
With our novel frame attention module, which fuses information across feature tracks, our tracker outperforms state-of-art methods on two datasets while being faster in terms of inference time.
Furthermore, our proposed method does not require intensive manual parameter tuning and can be adapted to new event cameras with our self-supervision strategy.
Ultimately, we can combine our event-based tracker with a KLT tracker to predict stable tracks in challenging scenarios.
This combination of standard and event cameras paves the path for the concept of sparingly triggering frames based on the tracking quality, which is a critical tool for future applications where runtime and power consumption are essential.
Finally, the proposed tracker seamlessly extends to sparse disparity estimation between an event and an image camera in a side-by-side setup, thereby providing the depth of each feature track.
% 

% \input{sections/supplementary}

% or


% \appendices

% \section{}

\ifCLASSOPTIONcompsoc
  \section*{Acknowledgments}
\else
  \section*{Acknowledgment}
\fi
The authors want to thank Javier Hidalgo-Carri\'{o} for the support of the EDS dataset.
This work was supported by the Swiss National Science Foundation through the National Centre of Competence in Research (NCCR) Robotics (grant number 51NF40\_185543), and the European Research Council (ERC) under grant agreement No. 864042 (AGILEFLIGHT).

\ifCLASSOPTIONcaptionsoff
  \newpage
\fi

% \newcolumntype{C}[1]{>{\centering}m{#1}}
\section*{\Large \bf Appendix: Data-driven Feature Tracking for Event Cameras with and without Frames}

\section{Future Work \& Limitations}
Since the EC and EDS datasets were recorded to benchmark pose estimation algorithms, they only contain static scenes.
Thus, we did not evaluate how our method, and especially our frame attention module performs in scenes with dynamic objects.
Nevertheless, we believe that our frame attention module can be useful for other trackers using event or standard cameras.
Finally, our method relies on the quality of the feature detection in grayscale images, which can suffer in challenging scenarios.
However, our self-supervision strategy opens up the possibility of also fine-tuning feature detectors for event cameras to increase the robustness of feature detection.

\subsection{Ablations}
We ablate each network block using a reference model without the frame attention module and grayscale inputs, see Tab.~\ref{tab:ablations}.
As verified by the performance drop (w/o augmentation), the augmentations during the training on synthetic data significantly boost the zero-shot transfer from synthetic to real-world data.
Furthermore, the recurrence in the feature encoder leads to longer feature age (w/recurrence), which is also achieved on a smaller scale by introducing the correlation map (w/o correlation).
While there is no improvement on the EC dataset, our proposed frame attention module significantly improves the performance on the challenging sequences of EDS.
This performance increase confirms the benefit of sharing information between similar feature tracks for challenging scenarios.
By adapting our network based on the frame attention module (Ref+Frame Attention) to real data using our self-supervision scheme, we achieve the highest tracking performance.
Finally, the frame attention module relies on state variables (w/o state) to fully exploit the potential of sharing information across features in a frame. 
Using a more complex LSTM for information propagation leads to overfitting, as shown by its lower End-Point-Error on simulated training data (0.569 vs. 0.720 for the default gating network).
\begin{table}[tb!]
\footnotesize
\vspace*{-8pt}
\caption{
Ablation experiments on the EDS and EC dataset.
}
\centering
\begin{tabular}{m{3.3cm}C{1.6cm}>{\centering\arraybackslash}m{1.6cm}}
     & \multicolumn{2}{c}{Expected FA $\uparrow$} \\
\cmidrule(lr){2-3}
Method  & EDS & EC \\
\hline
Reference Model             & 0.383 & 0.787 \\
\quad w/o correlation       & 0.341 & 0.684 \\
\quad w/o recurrence        & 0.301 & 0.606 \\
\quad w/o augmentation      & 0.178 & 0.599 \\
\hline
Ref + Frame Attention       & 0.451 & 0.787 \\
\quad w pose supervision    & 0.471 & 0.818 \\
\quad w/o state             & 0.385 & 0.791 \\
\quad w LSTM    & 0.423 & 0.661
\end{tabular}
\label{tab:ablations}
\vspace*{-10pt}
\end{table}

\section{Combination of Events and Frames}
\label{sec:combination_events_frames}
In a step to combine the contextual information of grayscale images and the high-latency information from events, we extended our event-based tracker using the popular KLT tracker for frames. 
Specifically, we use our event tracker to track features during the blind time between two frames and use the displacement prediction of our tracker as an initial guess for the KLT tracker once a new frame arrives.
This has the benefit of effectively mitigating the negative effects of large baselines between two frames caused by high-speed motion.
Additionally, the combination with our event tracker provides feature positions for the time in between two frames, significantly increasing the frequency of feature position updates.
On the other side, the KLT tracker can correct the feature position once reliable frame information is available. 
As used for the ground truth creation based on the camera poses, we use a KLT tracker with three hierarchical scales to cope with larger motion.
We compare the combination of our method and the KLT tracker (ours+KLT) against the pure KLT tracker for different pixel motions between frames, as reported in Fig.~\ref{fig:frameskip}
The different pixel motions are achieved by skipping frames in a sequence of the EC dataset, which corresponds to increasing the pixel motion between two frames.
As can be seen in Fig.~\ref{fig:frameskip}, the combination of ours and KLT performs comparably to a pure KLT tracker for small pixel displacement between frames.
However, with increasing pixel motion, the initial guess provided by our method helps the KLT tracker to track features over a longer time duration than a KLT tracker alone.
\begin{figure}[tb!]
\centering
\includegraphics[trim={12.5cm 0 0 0},clip,width=0.28\textwidth]{floats/images/frameskip.png}
\vspace*{-8pt}
\caption{
The tracking performance of KLT and our tracker combined with KLT (Ours+KLT) in relation to pixel motion per frame.
In combination, our event-based tracker can successfully help KLT predict larger displacement while KLT can refine the predictions of our tracker.
}

\label{fig:frameskip}
\end{figure} 

\section{Supervision}
\subsection{Synthetic Supervision}
\begin{figure}[t!]
\centering
\includegraphics[width=0.47\textwidth]{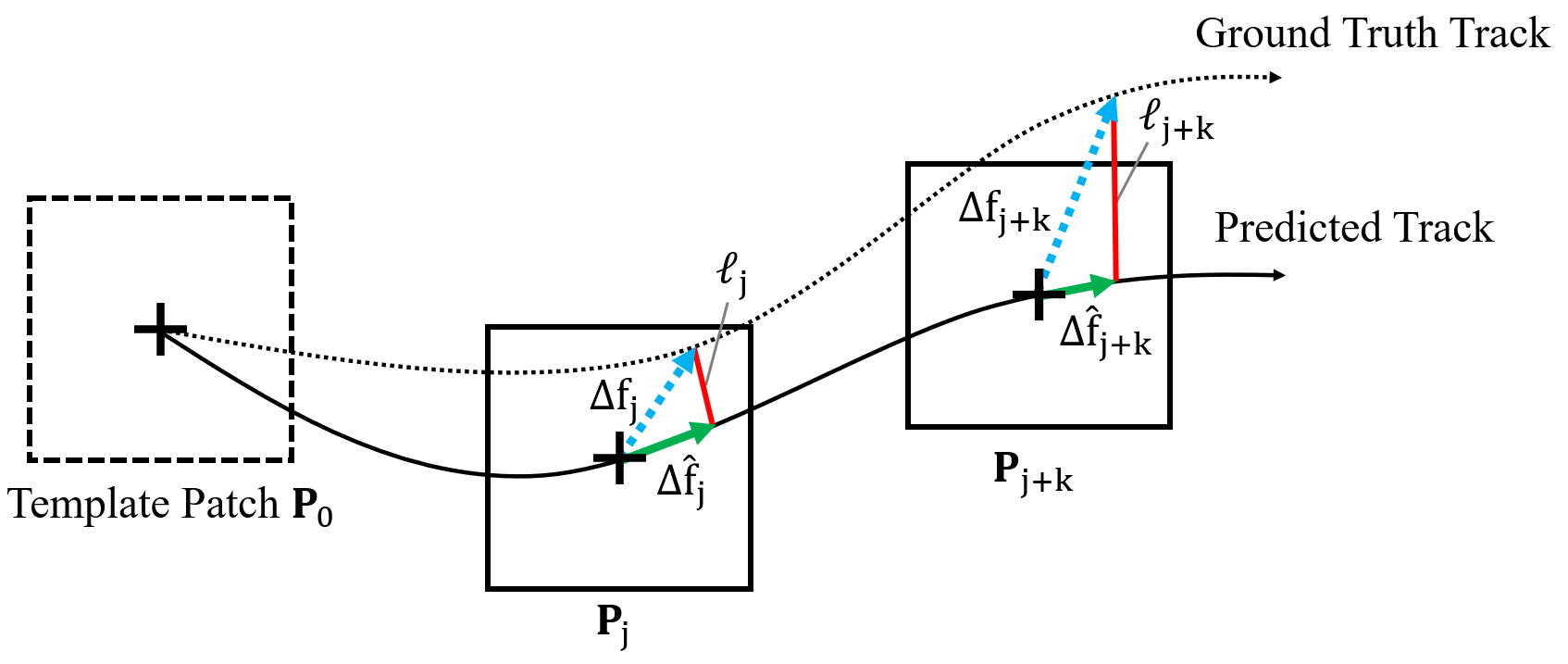}
\vspace{-12pt}
\caption{
The L1 distance $\ell_j$ between the predicted $\mathbf{\Delta \hat{f}_{j}}$ and the ground truth displacement $\mathbf{\Delta f_{j}}$ is used as a truncated loss, which is set to zero if the ground truth feature is outside of the current event patch $\mathbf{P}_j$, as shown for timestep $t_{j+k}$.
}
\label{fig:supervision_gt}
\vspace*{-8pt}
\end{figure}

On the Multiflow dataset, a loss based on the L1 distance can be directly applied for each prediction step $j$ between the predicted and ground truth relative displacement, see Fig.~\ref{fig:supervision_gt}. 
It is possible that the predicted feature tracks diverge beyond the template patch such that the next feature location is not in the current search.
Thus, if the difference between predicted and ground truth displacement $||\mathbf{\Delta \hat{f}_{j}} - \mathbf{\Delta f_{j}}||_1$ exceeds the patch radius $r$, we do not add the L1 distances to the final loss to avoid introducing noise in supervision.
Our truncated loss $\mathcal{L}_{rp}$ is formulated as follows.
\begin{gather}
    err_{j} = % 
  \begin{cases}
    ||\mathbf{\Delta \hat{f}_{j}} - \mathbf{\Delta f_{j}}||_1 \quad \quad & ||\mathbf{\Delta \hat{f}_{j}} - \mathbf{\Delta f_{j}}||_1 < r\\
    0 &\text{else}
  \end{cases}  \label{sup_loss_reproj_start}\\
    \mathcal{L}_{rp} = \frac{\sum_{j} \mathbbm{1}_{(err_{j} \neq 0)} err_{j}}{\sum_{j} \mathbbm{1}_{(err_{j} \neq 0)}} \label{sup_loss_reproj_end} 
\end{gather}

\subsection{Pose Supervision}
\label{sec:sup_pose_supervision}
To adapt the network to real events, we introduce a novel pose supervision loss solely based on ground truth poses of a calibrated camera.
The ground truth poses can easily be obtained for sparse timesteps $t_j$ using structure-from-motion algorithms, e.g., COLMAP~\cite{colmap_sfm}, or by an external motion capture system.
Since our supervision strategy relies on the triangulation of 3D points based on poses, it can only be applied in static scenes.
In the first step of the fine-tuning, our network predicts multiple feature tracks for one sequence. 
For each predicted track $i$, we compute the corresponding 3D point $\mathbf{X}_i$ using the direct linear transform~\cite{abdel2015direct}.
Specifically, for each feature location $\mathbf{x_j}$, we can write the projection equation assuming a pinhole camera model using the camera pose, represented as a rotation matrix $\mathbf{R}_{t_j}$ and a translation vector $\mathbf{T}_{t_j}$, at timestep $t_j$, and the calibration matrix $\mathbf{K}$, see Eq.~\ref{eq:proj}.
The resulting projection matrix can be expressed as matrix $\mathbf{M}_j$ consisting of column vectors $\mathbf{m^k}_{j}^T$ with $k \in \{1, 2, 3\}$.
\begin{equation}
        \mathbf{x}_j = \mathbf{K} \mathbf{[\mathbf{R}_{{t}_j} | \mathbf{T}_{{t}_j}]} \mathbf{X}_j = \mathbf{M}_j \mathbf{X}_j = \begin{bmatrix} \mathbf{m^1}_{j}^T  \\ \mathbf{m^2}_{j}^T \\ \mathbf{m^3}_{j}^T \end{bmatrix} \mathbf{X}_i \label{eq:proj} \\
\end{equation}
Using the direct linear transform, we can reformulate the projection equations as the homogenous linear system in Eq.~\ref{eq:homog_sys}.
By using SVD, we obtain the 3D point $\mathbf{X}_j$, which minimizes the least square error of Eq.~\ref{eq:homog_sys}.
\begin{gather}
    \begin{bmatrix} u_j \mathbf{m^3}_{j}^T -     \mathbf{m^2}_{j}^T  \\ 
                        \mathbf{m^1}_{j}^T - v_j \mathbf{m^3}_{j}^T  \\ 
                        ... 
    \end{bmatrix} = \mathbf{A} \mathbf{X}_i = 0  \label{eq:homog_sys}
\end{gather}
Once the 3D position of $\mathbf{X}_i$ is computed, we can find the reprojected pixel point $\mathbf{\Tilde{x}}_j$ for each timestep $t_j$ using perspective projection Eq.~\ref{eq:proj}.
The final pose supervision loss is then constructed based on the predicted feature $\mathbf{\hat{x}_{j}}$ and the reprojected feature $\mathbf{\Tilde{x}}_j$ for each available camera pose at timestep $t_j$.
As in the supervised setting of Eq.~\ref{sup_loss_reproj_end}, we use a truncated loss, which excludes the loss contribution if the reprojected feature is outside of the event patch.
\section{Dataset Split}
We use five sequences from the Event Camera dataset~\cite{Mueggler17ijrr} (EC) and four sequences from the Event-aided Direct Sparse Odometry dataset~\cite{Hidalgo2022cvpr} (EDS) as test sequences.
For fine-tuning, our pose supervision strategy is performed on five sequences from the EC and one sequence from the EDS dataset since EDS does not contain many sequences with ground truth pose in well-lit conditions.
The overview of the test and fine-tuning sequences is shown in Tab.~\ref{tab:sup_sequence}.
\begin{table}[ht!]
\caption{
Test and fine-tuning sequences for the EC and EDS dataset.
}
\centering
\begin{tabular}{m{0.3cm}m{1.1cm}m{3.1cm}m{2cm}}
 & Dataset  & Sequence Name & Frames \\
\hline
\multirow{9}{*}{\rotatebox[origin=c]{90}{Test}} & \multirow{5}{*}{EC}  & Shapes Translation & 8-88\\
                    & & Shapes Rotation & 165-245 \\
                    & & Shapes 6DOF & 485-485 \\
                    & & Boxes Translation & 330-410 \\
                    & & Boxes Rotation & 198-278 \\
\cmidrule(lr){2-4}
 & \multirow{4}{*}{EDS} & Peanuts Light & 160-386 \\
                    & & Rocket Earth Light & 338-438 \\
                    & & Ziggy In The Arena & 1350-1650 \\
                    & & Peanuts Running & 2360-2460 \\
\hline
\multirow{6}{*}{\rotatebox[origin=c]{90}{Fine-Tuning}} & \multirow{5}{*}{EC}  & boxes\_hdr & all \\
                    &     & calibration & all \\
                    &     & poster\_6dof & all \\
                    &     & poster\_rotation & all \\
                    &     & poster\_translation & all \\
\cmidrule(lr){2-4}
                    & EDS & all\_characters & all
\end{tabular}
\label{tab:sup_sequence}
\end{table}
For the disparity experiments on the M3ED dataset~\cite{chaney23cvpr}, we use in total of 33 sequences for training, 10 for validation, and 12 for testing.
The specific sequences are listed in Tab.~\ref{tab:sup_m3ed_sequence}.
\begin{table}[ht!]
\caption{
Training, Validation and Test split for the M3ED dataset.
}
\centering
\scalebox{0.70}{
\begin{tabular}{m{0.2cm}m{5cm}m{4.5cm}}
 & Sequence Name &  \\
\hline
\multirow{17}{*}{\rotatebox[origin=c]{90}{Training}}        & falcon\_indoor\_flight\_2 & car\_urban\_day\_penno\_small\_loop\\
                    & spot\_outdoor\_day\_srt\_under\_bridge\_2     & falcon\_forest\_into\_forest\_4 \\
                    & falcon\_forest\_road\_1                    & spot\_indoor\_stairwell \\
                    & falcon\_indoor\_flight\_1                  & spot\_outdoor\_day\_srt\_under\_bridge\_1 \\
                    & falcon\_outdoor\_day\_penno\_parking\_2      & spot\_outdoor\_day\_skatepark\_2 \\
                    & falcon\_forest\_into\_forest\_1             & falcon\_forest\_up\_down \\
                    & car\_urban\_day\_ucity\_small\_loop          & spot\_outdoor\_day\_art\_plaza\_loop \\
                    & car\_urban\_night\_city\_hall               & car\_urban\_day\_penno\_big\_loop \\
                    & falcon\_outdoor\_day\_penno\_parking\_1      & spot\_forest\_easy\_1 \\
                    & car\_urban\_night\_ucity\_small\_loop        & spot\_outdoor\_day\_skatepark\_1 \\
                    & falcon\_forest\_road\_2                    & car\_urban\_night\_rittenhouse \\
                    & falcon\_outdoor\_night\_penno\_parking\_1    & spot\_forest\_road\_1 \\
                    & car\_forest\_into\_ponds\_short             & car\_urban\_day\_rittenhouse \\
                    & car\_urban\_night\_penno\_small\_loop\_darker & car\_urban\_night\_penno\_small\_loop \\
                    & falcon\_outdoor\_night\_penno\_parking\_2    & spot\_forest\_hard \\
                    & car\_urban\_day\_city\_hall                 & car\_urban\_night\_penno\_big\_loop \\
                    & spot\_outdoor\_day\_rocky\_steps            &  \\
\hline
\multirow{5}{*}{\rotatebox[origin=c]{90}{Validation}} & falcon\_outdoor\_day\_fast\_flight\_1 & spot\_outdoor\_night\_penno\_short\_loop \\
                    & falcon\_indoor\_flight\_3       & spot\_outdoor\_day\_penno\_short\_loop \\
                    & spot\_indoor\_stairs            & car\_forest\_into\_ponds\_long \\
                    & falcon\_forest\_into\_forest\_2 & spot\_forest\_road\_3 \\
                    & spot\_indoor\_building\_loop    & spot\_forest\_easy\_2 \\
\hline
\multirow{5}{*}{\rotatebox[origin=c]{90}{Test}}             & car\_forest\_sand\_1 & falcon\_outdoor\_night\_high\_beams \\
                    & falcon\_forest\_road\_forest          & falcon\_outdoor\_day\_penno\_trees \\
                    & car\_forest\_tree\_tunnel             & spot\_indoor\_obstacles \\
                    & falcon\_outdoor\_day\_penno\_cars     & spot\_outdoor\_night\_penno\_plaza\_lights \\
                    & falcon\_outdoor\_day\_fast\_flight\_2 & car\_urban\_day\_horse \\
                    & spot\_outdoor\_day\_srt\_green\_loop  & falcon\_outdoor\_day\_penno\_plaza \\
\end{tabular}}
\label{tab:sup_m3ed_sequence}
\end{table}

\section{Multiflow Dataset}
\begin{figure*}[ht!]
\centering
\includegraphics[width=0.90\textwidth]{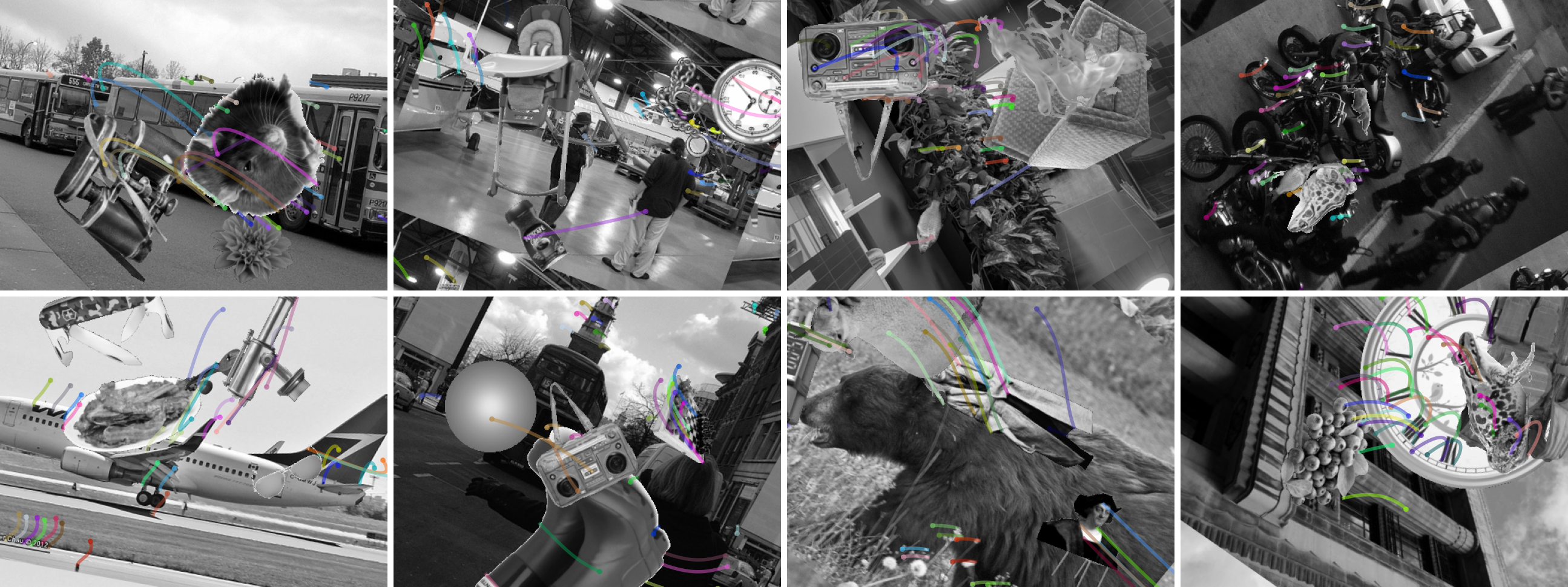}
\caption{
Samples from the Multiflow dataset including the ground truth tracks corresponding to extracted Harris features.
}
\label{fig:sup_multiflow}
\end{figure*}

To qualitatively show the gap between the simulated and the real data, we visualize in Fig.~\ref{fig:sup_multiflow} some examples from the Multiflow dataset~\cite{multiflow}, including the ground truth tracks corresponding to the extracted Harris features~\cite{Harris88}.
This sim-to-real gap can be reduced with our augmentation strategies on the Multiflow dataset and with our proposed fine-tuning strategy on real data, see Sec. 3.2.

\section{Network Architecture Details}
Tab.~\ref{tab:sup_network} shows the architectural details of our proposed network, which consists of a feature network and our proposed frame attention module.
In the first step, two patch encoders inside the feature network process the event and the grayscale patches, which have a patch size of 31 pixels.
After the correlation and the concatenation of the feature maps from both patch networks, a joint encoder refines the correlation map and introduces temporal information sharing through a ConvLSTM layer.
Finally, the frame attention module processes each feature in one frame using shared linear layers and one global multi-head attention over all features in a frame.
We refer to Fig. 2 in the main paper for the network overview.
\begin{table}[ht!]
\caption{
Network architecture. 
Each convolution layer is followed by LeakyReLU and BatchNorm layers whereas the linear layers are followed by LeakyReLu layers.
For the upsampling layers (Up), we use bilinear interpolation.
The three numbers after each convolution layer indicate the two kernel dimensions and the output channel dimension.
In the case of the linear layer, the single number stands for the output channels.
}
\centering
\begin{tabular}{m{0.3cm}m{3.7cm}m{2cm}}
 & Layer  & Spatial Size \\
\hline
\multirow{22}{*}{\rotatebox[origin=c]{90}{Feature Network (2$\times$Patch Encoders + Joint Encoder)}} & 2$\times$ Conv2D 1$\times$1$\times$32 & 31$\times$31 \\
                    & 2$\times$ Conv2D 5$\times$5$\times$64 & 23$\times$23 \\
                    & 2$\times$ Conv2D 5$\times$5$\times$128 & 15$\times$15 \\
                    & 2$\times$ Conv2D 3$\times$3$\times$256 & 5$\times$5 \\
                    & 2$\times$ Conv2D 1$\times$1$\times$384 & 1$\times$1 \\
                    & 2$\times$ Conv2D 1$\times$1$\times$384 & 1$\times$1 \\
                    & Up + Conv2D 1$\times$1$\times$384 & 5$\times$5 \\
                    & Conv2D 3$\times$3$\times$384 & 5$\times$5 \\
                    & Up + Conv2D 1$\times$1$\times$384 & 15$\times$15 \\
                    & Conv2D 3$\times$3$\times$384 & 15$\times$15 \\
                    & Up + Conv2D 1$\times$1$\times$384 & 23$\times$23 \\
                    & Conv2D 3$\times$3$\times$384 & 23$\times$23 \\
                    & Up + Conv2D 1$\times$1$\times$384 & 31$\times$31 \\
                    & Conv2D 3$\times$3$\times$384 & 31$\times$31  \\
                    & 2$\times$ Conv2D 3$\times$3$\times$384 & 31$\times$31  \\
                    \cmidrule(lr){2-3}
                    & Correlation Layer&  31$\times$31 \\
                    & 2$\times$ Conv2D 3$\times$3$\times$128 & 31$\times$31 \\
                    \cmidrule(lr){2-3}
                    & 2$\times$ Conv2D 3$\times$3$\times$64 & 15$\times$15 \\
                    & 2$\times$ Conv2D 3$\times$3$\times$128 & 7$\times$7 \\
                    & ConvLSTM 3$\times$3$\times$128 & 7$\times$7 \\
                    & 2$\times$ Conv2D 3$\times$3$\times$256 & 3$\times$3 \\
                    & Conv2D 3$\times$3$\times$256 & 1$\times$1 \\
                    \hline
\multirow{6}{*}{\rotatebox[origin=c]{90}{Frame Attention}} & Linear 256 & 1$\times$1 \\
                    & Linear 256 & 1$\times$1 \\
                    & MultiHead Attention & 1$\times$1 \\
                    & LayerScale 256 & 1$\times$1 \\
                    & Linear Gating 256 & 1$\times$1 \\
                    & Linear 2 & 1$\times$1 \\
\end{tabular}
\label{tab:sup_network}
\end{table}

\section{Quantitative Results \& Tracking Metrics}
As done in previous works~\cite{eklt, Alzugaray2020HASTE:Events}, we directly compare feature tracking metrics for a feature tracking methodology instead of computing pose errors using a pose estimation module.
While pose estimation is one application, it requires the tuning of many hyperparameters specifically for the tracker. 
Thus, it complicates evaluation and produces biased results. 
As tracking metrics, we report for each test sequence from the EC and EDS dataset the \textit{expected feature age} in Tab.~\ref{tab:sup_expected_fa}, the \textit{feature age} in Tab.~\ref{tab:sup_fa}, the \textit{inlier ratio} in Tab.~\ref{tab:sup_inlier_ratio} and the \textit{normalized tracking error} in Tab.~\ref{tab:sup_te}.
For the \textit{normalized tracking error}, we terminate the track if the distance to the ground truth exceeds 5 pixels, as done in~\cite{Alzugaray2020HASTE:Events}.
However, it is not obvious how to compute this metric if the tracking error is higher than 5 pixels directly after the initialization, as it occurred for the baseline methods in Tab.~\ref{tab:sup_te}.
Furthermore, this metric does not consider the duration of the predicted tracks, e.g., one feature can be tracked for a short time duration with a small tracking error, which would lead to a small normalized tracking error.
In contrast, a feature tracked for a long time horizon but with a higher distance to the ground truth will be assigned a higher tracking error.
This example shows that the \textit{normalized tracking error} on its own is not necessarily a good metric to evaluate stable and long feature tracks.
Thus, we decided to report the \textit{expected feature age} as a metric since it considers the tracking duration and the number of tracked features.
Moreover, the \textit{expected feature age} is computed over a range of termination thresholds with respect to the ground truth, which effectively eliminates this hyperparameter for the metric computation.
Specifically, the \textit{expected feature age} represents the multiplication of the \textit{normalized feature age} with the fraction of successfully predicted tracks over the number of given feature locations, defined as \textit{inlier ratio}.
A feature is defined to be tracked successfully if the predicted feature location at the second timestep after initialization is in the termination threshold to the ground truth location.
The \textit{normalized feature age} is computed for the successfully tracked features based on the division of the time duration until the predicted feature exceeds the termination threshold to the ground truth location by the duration of the ground truth tracks.
Because of the range of termination thresholds and the consideration of the number of successfully tracked features, the \textit{expected feature age} represents an expressive and objective metric for reporting the tracking performance.
Compared to~\cite{Manderscheid19cvpr}, we evaluate the tracking performance and thus use the same features for each method. 
Furthermore, our evaluation focuses on the introduced Expected Feature Age to account for the impact of outliers, which is typically ignored. 

\begin{table*}[ht!]
\caption{
The performance of our proposed and the baseline trackers on the EDS and EC dataset in terms of \textit{Expected Feature Age}.
}
\centering
\begin{tabular}{m{3.1cm}C{2cm}C{2cm}C{2cm}C{2cm}>{\centering\arraybackslash}m{2cm}}
     & \multicolumn{5}{c}{Expected FA $\uparrow$}  \\
\cmidrule(lr){2-6}
Sequence  & ICP~\cite{Kueng16iros}  & EM-ICP~\cite{Zhu17icra} & HASTE~\cite{Alzugaray2020HASTE:Events} & EKLT~\cite{eklt} & \textbf{Ours}  \\
\hline
Shapes Translation      & 0.306 & 0.402 & 0.564 & 0.740 & 0.856 \\
Shapes Rotation         & 0.339 & 0.320 & 0.582 & 0.806 & 0.793 \\
Shapes 6DOF             & 0.129 & 0.242 & 0.043 & 0.696 & 0.882 \\
Boxes Translation       & 0.261 & 0.354 & 0.368 & 0.644 & 0.869 \\
Boxes Rotation          & 0.188 & 0.349 & 0.447 & 0.865 & 0.691 \\
\hline
EC Avg                  & 0.245 & 0.334 & 0.427 & 0.775 & 0.818 \\
\hline
Peanuts Light           & 0.044 & 0.077 & 0.076 & 0.260 & 0.420 \\
Rocket Earth Light      & 0.045 & 0.158 & 0.085 & 0.175 & 0.291 \\
Ziggy In The Arena      & 0.039 & 0.149 & 0.057 & 0.231 & 0.746 \\
Peanuts Running         & 0.028 & 0.095 & 0.033 & 0.153 & 0.428 \\
\hline
EDS Avg                 & 0.040 & 0.120 & 0.063 & 0.205 & 0.472 \\
\hline
\end{tabular}
\label{tab:sup_expected_fa}
\vspace*{15pt}
\end{table*}

\begin{table*}[ht!]
\caption{
The performance of our proposed and the baseline trackers on the EDS and EC dataset in terms of \textit{Feature Age FA}.
}
\centering
\begin{tabular}{m{3.1cm}C{2cm}C{2cm}C{2cm}C{2cm}>{\centering\arraybackslash}m{2cm}}
     & \multicolumn{5}{c}{Feature Age (FA) $\uparrow$}  \\
\cmidrule(lr){2-6}
Sequence  & ICP~\cite{Kueng16iros}  & EM-ICP~\cite{Zhu17icra} & HASTE~\cite{Alzugaray2020HASTE:Events} & EKLT~\cite{eklt} & \textbf{Ours}  \\
\hline
Shapes Translation      & 0.307 & 0.403 & 0.589 & 0.839 & 0.861 \\
Shapes Rotation         & 0.341 & 0.320 & 0.613 & 0.833 & 0.797 \\
Shapes 6DOF             & 0.169 & 0.248 & 0.133 & 0.817 & 0.899 \\
Boxes Translation       & 0.268 & 0.355 & 0.382 & 0.682 & 0.872 \\
Boxes Rotation          & 0.191 & 0.356 & 0.492 & 0.883 & 0.695 \\
\hline
EC Avg                  & 0.256 & 0.337 & 0.442 & 0.811 & 0.825 \\
\hline
Peanuts Light           & 0.050 & 0.084 & 0.086 & 0.284 & 0.447 \\
Rocket Earth Light      & 0.103 & 0.298 & 0.162 & 0.425 & 0.648 \\
Ziggy In The Arena      & 0.043 & 0.153 & 0.082 & 0.419 & 0.748 \\
Peanuts Running         & 0.043 & 0.108 & 0.054 & 0.171 & 0.460 \\
\hline
EDS Avg                 & 0.060 & 0.161 & 0.096 & 0.325 & 0.576 \\
\hline
\end{tabular}
\label{tab:sup_fa}
\vspace*{15pt}
\end{table*}

\begin{table*}[ht!]
\caption{
The performance of our proposed and the baseline trackers on the EDS and EC dataset in terms of \textit{Inlier Ratio}.
}
\centering
\begin{tabular}{m{3.1cm}C{2cm}C{2cm}C{2cm}C{2cm}>{\centering\arraybackslash}m{2cm}}
     & \multicolumn{5}{c}{Inlier Ratio $\uparrow$}  \\
\cmidrule(lr){2-6}
Sequence  & ICP~\cite{Kueng16iros}  & EM-ICP~\cite{Zhu17icra} & HASTE~\cite{Alzugaray2020HASTE:Events} & EKLT~\cite{eklt} & \textbf{Ours}  \\
\hline
Shapes Translation      & 0.986 & 0.916 & 0.957 & 0.882 & 0.962 \\
Shapes Rotation         & 0.962 & 0.955 & 0.950 & 0.968 & 0.950 \\
Shapes 6DOF             & 0.696 & 0.755 & 0.325 & 0.852 & 0.946 \\
Boxes Translation       & 0.937 & 0.937 & 0.963 & 0.945 & 0.980 \\
Boxes Rotation          & 0.946 & 0.798 & 0.908 & 0.980 & 0.949 \\
\hline
EC Avg                  & 0.905 & 0.872 & 0.820 & 0.925 & 0.957 \\
\hline
Peanuts Light           & 0.740 & 0.868 & 0.815 & 0.780 & 0.802 \\
Rocket Earth Light      & 0.369 & 0.401 & 0.293 & 0.375 & 0.374 \\
Ziggy In The Arena      & 0.421 & 0.884 & 0.609 & 0.469 & 0.927 \\
Peanuts Running         & 0.502 & 0.578 & 0.531 & 0.700 & 0.750 \\
\hline
EDS Avg                 & 0.508 & 0.683 & 0.562 & 0.581 & 0.713 \\
\hline
\end{tabular}
\label{tab:sup_inlier_ratio}
\vspace*{15pt}
\end{table*}

\begin{table*}[ht!]
\caption{
The performance of our proposed and the baseline trackers on the EDS and EC dataset in terms of \textit{Track Normalized Error}.
}
\centering
\begin{tabular}{m{3.1cm}C{2cm}C{2cm}C{2cm}C{2cm}>{\centering\arraybackslash}m{2cm}}
     & \multicolumn{5}{c}{Track Normalized Error $\downarrow$}  \\
\cmidrule(lr){2-6}
Sequence  & ICP~\cite{Kueng16iros}  & EM-ICP~\cite{Zhu17icra} & HASTE~\cite{Alzugaray2020HASTE:Events} & EKLT~\cite{eklt} & \textbf{Ours}  \\
\hline
Shapes Translation      & 1.943 & 3.941 & 2.628 & 1.104 & 1.153 \\
Shapes Rotation         & 1.870 & 2.614 & 2.536 & 1.723 & 1.981 \\
Shapes 6DOF             &   -   &   -   &   -   & 1.833 & 1.702 \\
Boxes Translation       & 2.289 & 2.613 & 2.109 & 1.227 & 1.166 \\
Boxes Rotation          & 2.571 & 3.855 & 3.383 & 1.375 & 1.836 \\
\hline
EC Avg                  & 2.168 & 3.256 & 2.664 & 1.452 & 1.568 \\
\hline
Peanuts Light           & 3.185 & 2.323 & 2.432 & 3.560 & 3.957 \\
Rocket Earth Light      &   -   & 4.062 &   -   & 2.405 & 3.599 \\
Ziggy In The Arena      &   -   & 3.407 & 2.672 &   -   & 2.673 \\
Peanuts Running         &   -   &   -   &   -   & 3.812 & 3.444 \\
\hline
EDS Avg                 & 3.185 & 3.264 & 2.552 & 3.259 & 3.418 \\
\hline
\end{tabular}
\label{tab:sup_te}
\vspace*{15pt}
\end{table*}

\begin{table*}[ht!]
\caption{
The performance of the \textit{reference model} when trained with different input event representations.
}
\centering
\begin{tabular}{m{3.1cm}C{2cm}C{2cm}C{2cm}>{\centering\arraybackslash}m{2cm}}
     & \multicolumn{4}{c}{Expected FA $\uparrow$}  \\
\cmidrule(lr){2-5}
Sequence  & SBT-Max  & SBT No Norm & SBT~\cite{sbt} & Voxel Grids~\cite{voxel_grids}  \\
\hline
Shapes Translation                          & 0.780   & 0.160                                                 & 0.887 & 0.802       \\
Shapes Rotation                             & 0.747   & 0.057                                                 & 0.823 & 0.799       \\
Shapes 6DOF                                 & 0.881   & 0.006                                                 & 0.882 & 0.882       \\
Boxes Translation                           & 0.849   & 0.160                                                 & 0.831 & 0.769       \\
Boxes Rotation                              & 0.614   & 0.057                                                 & 0.677 & 0.638       \\ \hline
EC Avg                                      & 0.774   & 0.088                                                 & 0.820 & 0.778       \\ \hline
Peanuts Light                               & 0.388   & 0.020                                                 & 0.373 & 0.372       \\
Rocket Earth Light                          & 0.271   & 0.009                                                 & 0.284 & 0.282       \\
Ziggy In The Arena                          & 0.686   & 0.040                                                 & 0.708 & 0.694       \\
Peanuts Running                             & 0.059   & 0.024                                                 & 0.073 & 0.150       \\ \hline
EDS Avg                                     & 0.351   & 0.023                                                 & 0.359 & 0.374       \\ \hline
\end{tabular}
\label{tab:sup_ablate_input}
\vspace*{15pt}
\end{table*}

\begin{table*}[ht!]
\caption{
The performance of the \textit{reference model} when trained with different augmentation parameters.
}
\centering
\begin{tabular}{m{3.1cm}C{2cm}C{2cm}C{2cm}>{\centering\arraybackslash}m{2cm}}
     & \multicolumn{4}{c}{Expected FA $\uparrow$}  \\
\cmidrule(lr){2-5}
Sequence  & R15 S10 T3  & R30 & T5 & No Aug  \\
\hline
Shapes Translation                          & 0.691   & 0.861                                                 & 0.720 & 0.723         \\
Shapes Rotation                             & 0.726   & 0.766                                                 & 0.697 & 0.617         \\
Shapes 6DOF                                 & 0.883   & 0.882                                                 & 0.876 & 0.499         \\
Boxes Translation                           & 0.809   & 0.791                                                 & 0.743 & 0.501         \\
Boxes Rotation                              & 0.616   & 0.703                                                 & 0.448 & 0.337         \\ \hline
EC Avg                                      & 0.745   & 0.801                                                 & 0.697 & 0.535         \\ \hline
Peanuts Light                               & 0.361   & 0.384                                                 & 0.337 & 0.311         \\
Rocket Earth Light                          & 0.284   & 0.275                                                 & 0.274 & 0.094         \\
Ziggy In The Arena                          & 0.658   & 0.699                                                 & 0.669 & 0.166         \\
Peanuts Running                             & 0.080   & 0.098                                                 & 0.156 & 0.028         \\ \hline
EDS Avg                                     & 0.346   & 0.364                                                 & 0.359 & 0.150         \\ \hline
\end{tabular}
\label{tab:sup_ablate_augm}
\end{table*}
\begin{figure*}[ht!]
\centering
\includegraphics[width=0.80\textwidth]{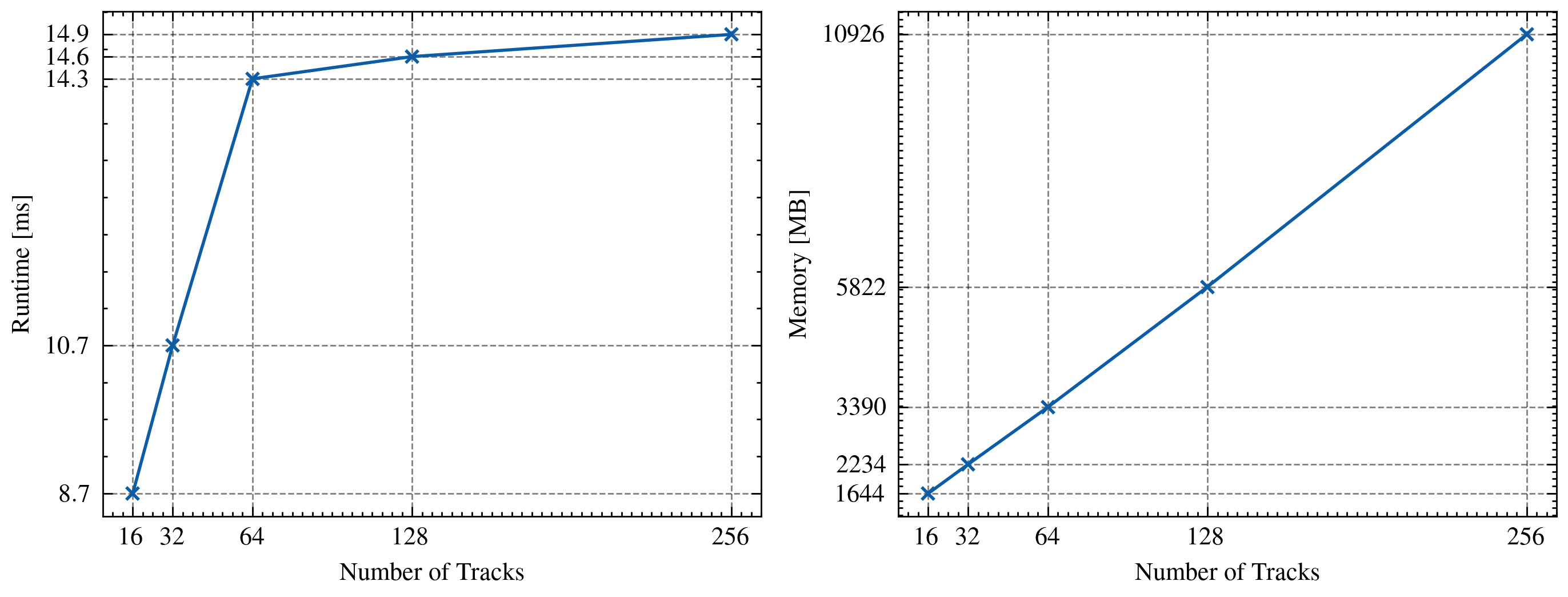}
\caption{
\textbf{Runtime and Memory for Sparse Disparity Estmation}
The runtime and memory used by our method for different numbers of processed feature tracks.
}
\label{fig:sup_tracks_runtime}
\end{figure*} 
\begin{figure*}[ht!]
\centering
\includegraphics[width=0.80\textwidth]{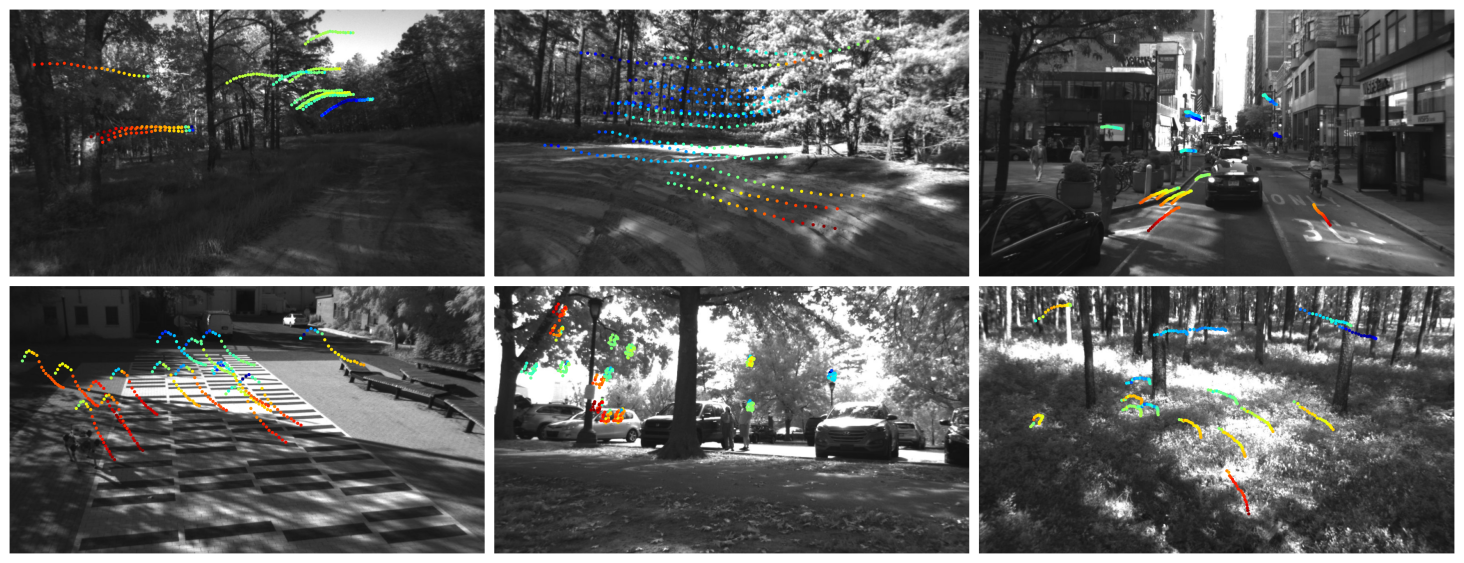}
\caption{
\textbf{Predicted Disparities}
The tracks on the M3ED dataset are colorized based on their corresponding disparity values, ranging from blue to red to represent small to large disparity values predicted by our method.
}
\label{fig:sup_qualitative_disparity}
\end{figure*}

\section{Input Event Representation}
For the aligned grayscale and event tracker setting, our method uses spatially and temporally aligned frames and events similar to previous works~\cite{eklt}.
This data can be recorded by cameras outputting directly events and images with one sensor (ATIS) or with beam splitter setups using two cameras aligned through a mirror setup.
To provide the events in a patch as input to our network, we first convert them to a dense event representation.
Specifically, we use a maximal timestamp version of SBT~\cite{sbt}, named SBT-Max, which consists of five temporal bins for positive and negative polarity leading to 10 channels.
Because of these design choices, the used event representation can be considered a combination between TimeSurface~\cite{Mueggler17bmvc} and SBT~\cite{sbt}.
In each temporal bin, we assign to each pixel coordinate the relative timestamp of the most recent event during the time interval of the temporal bin.
For the EC and EDS dataset, we convert events inside a \unit[10]{ms} and \unit[5]{ms} window, respectively.

\section{Additional Ablation Experiments}
We ablated the event input representation as well as the augmentation parameters used during training.
Due to time reasons, we performed the following ablation experiments by training the \textit{reference model}, which does not include the frame attention module, for 70000 steps instead of 140000.

\subsection{Input Representations}
The input event representation to an event-based network is an important consideration. Ideally, we aim to preserve as much of the spatiotemporal information as possible while minimizing the computational overhead of representation generations.
We train the reference network with different representations: voxel grids~\cite{voxel_grids}, Stacking Based on Time (SBT)~\cite{sbt}, a non-normalized version of SBT (SBTNo Norm) and a maximal timestamp version of SBT we call SBT-Max where each pixel is assigned the timestamp of the most recent event.
The results are shown in Tab.~\ref{tab:sup_ablate_input}. 
While many event-based networks have demonstrated promising results with voxel grids, their interpolation-based construction is computationally expensive. 
In contrast, SBT is a simpler, synchronous event representation that is more efficient. Each pixel simply accumulates or "stacks" incoming events.
We find that SBT achieves competitive \textit{Expected FA} compared to voxel grids on nearly all sequences. 
However, the performance of SBT degrades significantly without normalizing based on the number of events in the frame. 
In contrast to normalizing by the number of events, SBT-Max is normalized using the duration of the time window. 
In practice, the statistic-free normalization procedure of SBT-Max means that events outside the neighborhoods of tracked features can be ignored.
Because of this deployment advantage and the competitive performance despite its more simplistic normalization, we select SBT-Max as event representation.

\subsection{Augmentation Parameters}
To validate the utility of our augmentation strategy, we train the reference network with different augmentation parameters.
In Tab.~\ref{tab:sup_ablate_augm}, we present the experimental results for using rotations (R) of up to $\pm 30^\circ$, scaling (S) of up to $\pm 10\%$, and translations (T) of up to $\pm 5px$.
The default training settings use rotations of up to $\pm 15^\circ$, scaling of up to $\pm 10\%$, and translations of up to $\pm 3px$.
Without augmentation, we observe significant degradation on both datasets. 
The benefit of additional translation augmentation is inconclusive, given the degradation on EC and improvement on EDS. 
Lastly, with increased rotation augmentation, we observe that the performance improves on average for both datasets.
Notably, the sim-to-real gap can be demonstrated when the reference network is trained without on-the-fly augmentations (No Aug).
As a consequence, the expected FA metric on the real dataset drops considerably, from 0.745 to 0.535 on the EC dataset and from 0.346 to 0.150 on the EDS dataset.
The sim-to-real gap is also apparent in the qualitative comparisons between the multiflow dataset (Fig.~\ref{fig:sup_multiflow}) and the real datasets, EDS and EC (Fig. 5 in the main manuscript).

\section{Runtime and Memory Study}
For the disparity estimation task, we evaluate the impact of processing different numbers of feature tracks on the runtime and memory used by our method.
As shown in Fig.~\ref{fig:sup_tracks_runtime}, the runtime increases fast when processing 64 instead of 16 feature tracks.
However, increasing the number of feature tracks from 64 to 256 adds only 0.6 ms to the runtime, which can be explained by the parallelization capability of a GPU.
This confirms that even when processing 256 feature tracks, our proposed method achieves a lower runtime than the closest dense baseline (14.9 ms vs. 25.2 ms).
Nevertheless, we did observe a less favorable linear increase in GPU memory usage, reaching 10.9 GB when processing 256 feature tracks.
The memory usage can potentially be optimized by applying the first two U-Net modules in the feature network to the entire frame and event representation before extracting patches to reduce redundant computations of overlapping patches.
It is also worth noting that our tests were conducted using PyTorch, which is not optimized for inference speed. 
Deploying the method directly in C++ or TensorRT would likely lead to significantly faster and more efficient performance.
Overall, the proposed sparse disparity estimation setting trades off access to global frame information for faster, more efficient computation. 
Although our method does not have access to full frames, our proposed frame attention module enables the sharing of global frame information while accumulating temporal information along feature tracks.
As a result, our method achieves performance comparable to dense approaches while benefiting from the efficiency of sparse disparity estimation.

\section{Qualitative Disparity Predictions}
Fig~\ref{fig:sup_qualitative_disparity} shows disparity tracks predicted with our method on the M3ED dataset featuring different platforms (Quadrotor, Quadruped, Car).
\newpage



%\bibliographystyle{IEEEtran}
% \end{thebibliography}

\bibliographystyle{IEEEtran}
\bibliography{all}


% \end{IEEEbiography}


% % insert where needed to balance the two columns on the last page with
% %\newpage



\begin{IEEEbiography}[{\includegraphics[width=1in,height=1.25in,clip,keepaspectratio]{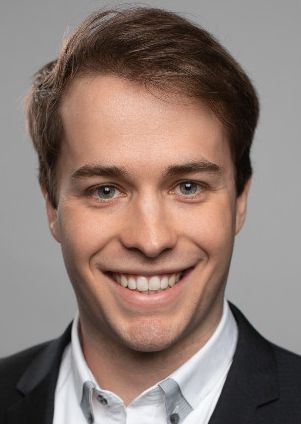}}]{Nico Messikommer} %
holds an M.Sc. in Robotics, Systems, and Control from ETH Zürich, Switzerland, which he obtained in 2019 following his B.Sc. in Mechanical Engineering in 2017. Currently pursuing a Ph.D. at the University of Zurich under the supervision of Prof. Davide Scaramuzza, his research centers on the intersection of computer vision and robotics, exploring topics such as unsupervised domain adaptation, event cameras, and reinforcement learning. His contributions were recognized by the nomination as an Award Candidate at the 2023 Conference on Computer Vision and Pattern Recognition (CVPR).
\end{IEEEbiography}
\vskip -2.6\baselineskip plus -1fil

\begin{IEEEbiography}[{\includegraphics[width=1in,height=1.25in,clip,keepaspectratio]{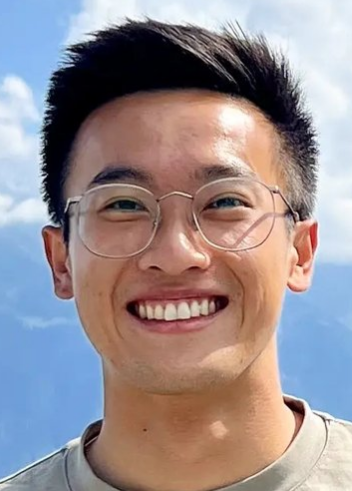}}]{Carter Fang} %
Carter Fang holds an M.Sc. in Robotics, Systems, and Control from ETH Zürich, Switzerland, which he obtained in 2022 after his B.Sc. in Mechanical Engineering obtained in 2019. His Master thesis completed under the guidance of Prof. Davide Scaramuzza was nominated for the Best Paper Award at the 2023 Conference on Computer Vision and Pattern Recognition (CVPR). He is currently a Research Engineer at Waabi working on Multimodal Perception and Prediction for autonomous freight under the guidance of Prof. Raquel Urtasun. 
\end{IEEEbiography}
\vskip -2.6\baselineskip plus -1fil

\begin{IEEEbiography}[{\includegraphics[width=1in,height=1.25in,clip,keepaspectratio]{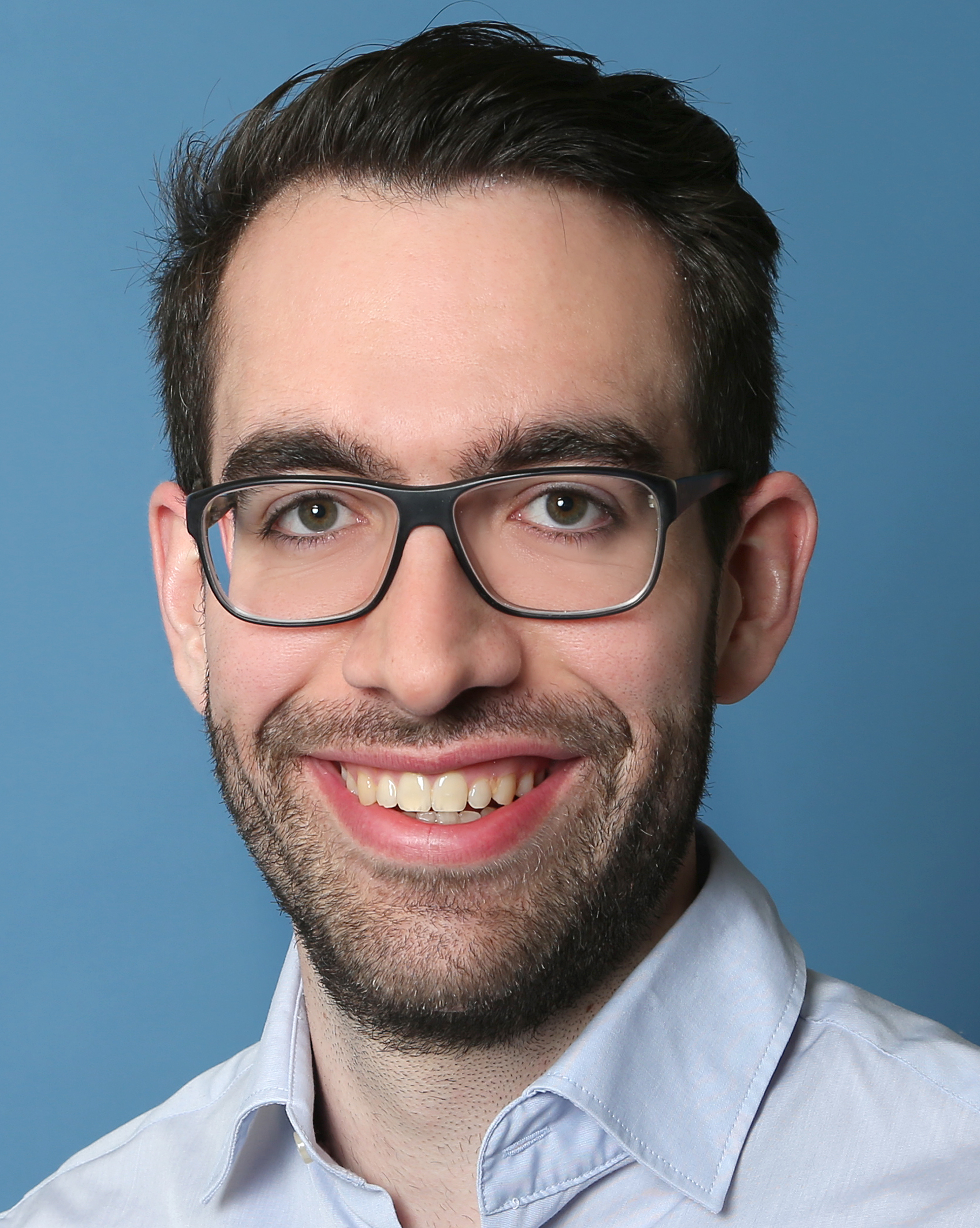}}]{Mathias Gehrig} %
obtained his M.Sc. in Robotics, Systems and Control from ETH Zürich, Switzerland, in 2016, after receiving his B.Sc. in Mechanical Engineering in 2013.
As a Ph.D. candidate in computer science at the University of Zurich, supervised by Prof. Davide Scaramuzza, he focuses on the application of machine learning for real-time computer vision and robotics.
Notably, his work was nominated for the Best Paper Award at the 2023 Conference on Computer Vision and Pattern Recognition (CVPR).
\end{IEEEbiography}
\vskip -2.6\baselineskip plus -1fil

\begin{IEEEbiography}[{\includegraphics[width=1in,height=1.25in,clip,keepaspectratio]{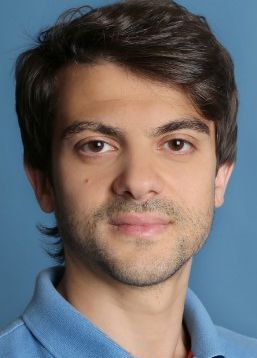}}]{Giovanni Cioffi} %
holds an M.Sc. in Mechanical Engineering from ETH Zürich, Switzerland, which he obtained in 2019. He is currently pursuing a Ph.D. at the University of Zürich under the supervision of Prof. Davide Scaramuzza. His research centers on the intersection of computer vision and robotics, exploring topics such as visual(-inertial) odometry and SLAM. His contributions were recognized by multiple awards in top-tier robotic conferences and journals, such as the IROS 2023 Best Paper Award and the RA-L 2021 Best Paper Award.
\end{IEEEbiography}
\vskip -2.6\baselineskip plus -1fil

\begin{IEEEbiography}[{\includegraphics[width=1in,height=1.25in,clip,keepaspectratio]{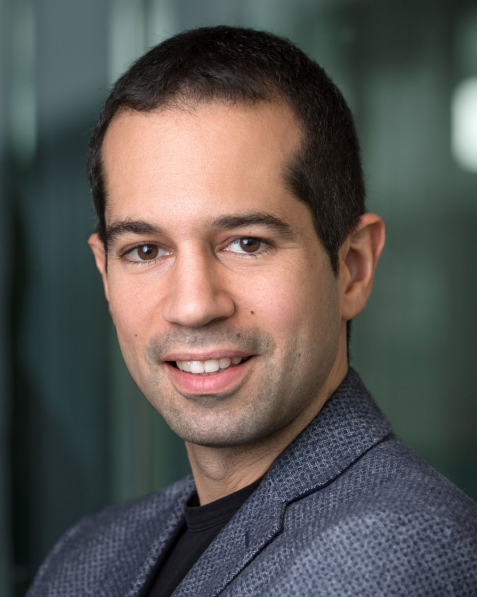}}]{Davide Scaramuzza} %
is a Professor of Robotics and Perception at the University of Zurich. He did his Ph.D. at ETH Zurich, a postdoc at the University of Pennsylvania, and was a visiting professor at Stanford University. His research focuses on autonomous, agile microdrone navigation using standard and event-based cameras. He pioneered autonomous, vision-based navigation of drones, which inspired the navigation algorithm of the NASA Mars helicopter and many drone companies. He contributed significantly to visual-inertial state estimation, vision-based agile navigation of microdrones, and low-latency, robust perception with event cameras, which were transferred to many products, from drones to automobiles, cameras, AR/VR headsets, and mobile devices. In 2022, his team demonstrated that an AI-controlled, vision-based drone could outperform the world champions of drone racing, a result that was published in Nature. He is a consultant for the United Nations on disaster response and disarmament. He has won many awards, including an IEEE Technical Field Award, the IEEE Robotics and Automation Society Early Career Award, a European Research Council Consolidator Grant, a Google Research Award, two NASA TechBrief Awards, and many paper awards. In 2015, he co-founded Zurich-Eye, today Meta Zurich, which developed the world-leading virtual-reality headset Meta Quest. In 2020, he co-founded SUIND, which builds autonomous drones for precision agriculture. Many aspects of his research have been featured in the media, such as The New York Times, The Economist, and Forbes.
\end{IEEEbiography}
\vskip -2.6\baselineskip plus -1fil


%\vfill


\end{document}